\def\year{2019}\relax
\pdfoutput=1
\documentclass[letterpaper]{article}
\usepackage[pdfauthor={Sabrina J. Mielke, Jason Eisner},
pdftitle={Spell Once, Summon Anywhere: A Two-Level Open-Vocabulary Language Model},
breaklinks=true, colorlinks=false, pdfborder={0 0 0}, hidelinks]{hyperref}
\usepackage{aaai19}
\usepackage{times}
\usepackage{helvet}
\usepackage{courier}
\usepackage{url}
\usepackage{graphicx}
\newcommand{\citet}[1]{\citeauthor{#1}~\shortcite{#1}}
\newcommand{\citep}{\cite}

\usepackage{latexsym}

\usepackage{amsmath}
\usepackage{amssymb}
\usepackage{appendix}

\usepackage{xspace}

\usepackage{mathptmx}
\DeclareSymbolFont{gregasM}{OT1}{cmr}{m}{n}
\DeclareMathSymbol{\Gamma}{\mathalpha}{gregasM}{0}
\DeclareMathSymbol{\Delta}{\mathalpha}{gregasM}{1}
\DeclareMathSymbol{\Theta}{\mathalpha}{gregasM}{2}
\DeclareMathSymbol{\Lambda}{\mathalpha}{gregasM}{3}
\DeclareMathSymbol{\Xi}{\mathalpha}{gregasM}{4}
\DeclareMathSymbol{\Pi}{\mathalpha}{gregasM}{5}
\DeclareMathSymbol{\Sigma}{\mathalpha}{gregasM}{6}
\DeclareMathSymbol{\Upsilon}{\mathalpha}{gregasM}{7}
\DeclareMathSymbol{\Phi}{\mathalpha}{gregasM}{8}
\DeclareMathSymbol{\Psi}{\mathalpha}{gregasM}{9}
\DeclareMathSymbol{\Omega}{\mathalpha}{gregasM}{10}
\DeclareSymbolFont{gregasm}{OML}{cmm}{m}{it}
\DeclareMathSymbol{\alpha}{\mathalpha}{gregasm}{11}
\DeclareMathSymbol{\beta}{\mathalpha}{gregasm}{12}
\DeclareMathSymbol{\gamma}{\mathalpha}{gregasm}{13}
\DeclareMathSymbol{\delta}{\mathalpha}{gregasm}{14}
\DeclareMathSymbol{\epsilon}{\mathalpha}{gregasm}{15}
\DeclareMathSymbol{\zeta}{\mathalpha}{gregasm}{16}
\DeclareMathSymbol{\eta}{\mathalpha}{gregasm}{17}
\DeclareMathSymbol{\theta}{\mathalpha}{gregasm}{18}
\DeclareMathSymbol{\iota}{\mathalpha}{gregasm}{19}
\DeclareMathSymbol{\kappa}{\mathalpha}{gregasm}{20}
\DeclareMathSymbol{\lambda}{\mathalpha}{gregasm}{21}
\DeclareMathSymbol{\mu}{\mathalpha}{gregasm}{22}
\DeclareMathSymbol{\nu}{\mathalpha}{gregasm}{23}
\DeclareMathSymbol{\xi}{\mathalpha}{gregasm}{24}
\DeclareMathSymbol{\pi}{\mathalpha}{gregasm}{25}
\DeclareMathSymbol{\rho}{\mathalpha}{gregasm}{26}
\DeclareMathSymbol{\sigma}{\mathalpha}{gregasm}{27}
\DeclareMathSymbol{\tau}{\mathalpha}{gregasm}{28}
\DeclareMathSymbol{\upsilon}{\mathalpha}{gregasm}{29}
\DeclareMathSymbol{\phi}{\mathalpha}{gregasm}{30}
\DeclareMathSymbol{\chi}{\mathalpha}{gregasm}{31}
\DeclareMathSymbol{\psi}{\mathalpha}{gregasm}{32}
\DeclareMathSymbol{\omega}{\mathalpha}{gregasm}{33}
\DeclareMathSymbol{\varepsilon}{\mathalpha}{gregasm}{34}
\DeclareMathSymbol{\vartheta}{\mathalpha}{gregasm}{35}
\DeclareMathSymbol{\varpi}{\mathalpha}{gregasm}{36}
\DeclareMathSymbol{\varrho}{\mathalpha}{gregasm}{37}
\DeclareMathSymbol{\varphi}{\mathalpha}{gregasm}{39}

\DeclareMathAlphabet{\mathcal}{OMS}{cmsy}{m}{n}

\usepackage{ocr}
\usepackage{microtype}
\newcommand{\charword}[1]{\scalebox{.9}{\ocrfamily #1}}
\newcommand{\spelling}{s}
\newcommand{\wughighlight}[1]{{\ocrfamily\color{black} #1}}

\usepackage{tikz}
\usetikzlibrary{arrows,matrix,positioning,fit,calc,shapes,decorations.pathreplacing,svg.path}
\pgfdeclarelayer{background}
\pgfdeclarelayer{backgroundoverlay}
\pgfsetlayers{background,backgroundoverlay,main}
\usepackage{pgfplots}
\pgfplotsset{compat=1.13}
\usepackage{adjustbox}

\usepackage{listings}
\lstdefinestyle{default}
{ basicstyle=\small\ttfamily
, identifierstyle=
, stringstyle=
, keywordstyle=\bfseries
, captionpos=b 
, columns=fullflexible
, keepspaces=true
, sensitive=true
, showstringspaces=false
, firstnumber=auto
, showspaces=false
, showtabs=false
, xrightmargin=6.5em
, xleftmargin=6.5em
, numbersep=0.8em
, framerule=0.5pt
, rulesep=-2.4em
, tabsize=2
, frame=trbL
, backgroundcolor=\color{gray!5}
, rulecolor=\color{black!20}
}
\usepackage[super]{nth}
\usepackage{varwidth}

\newcommand{\circnum}[1]{\raisebox{.5pt}{\textcircled{\raisebox{-.5pt}{\scalebox{.8}{\(#1\)}}}}}
\newcommand{\ofw}[1]{\ensuremath{{#1}\raisebox{.06em}{\scalebox{.7}{$(w)$}}}}

\newcommand{\overflowsamples}[1]{}

\usepackage{multirow}

\usepackage[capitalise]{cleveref}

\crefname{subsection}{\S\!}{\S\S\!}
\crefformat{section}{\S#2#1#3}

\title{Spell Once, Summon Anywhere: \\ A Two-Level Open-Vocabulary Language Model}

\author{
    Sabrina J. Mielke \and Jason Eisner\\
    Department of Computer Science, Johns Hopkins University, Baltimore, MD, USA\\
    {\tt sjmielke@jhu.edu}, {\tt jason@cs.jhu.edu}  \\}

\date{}

\begin{document}
\maketitle

\thispagestyle{plain}
\pagestyle{plain}

\begin{abstract}
We show how the spellings of known words can help us deal with unknown words in open-vocabulary NLP tasks.
The method we propose can be used to extend any closed-vocabulary generative model, but in this paper we specifically consider the case of neural language modeling.
Our Bayesian generative story combines a standard RNN language model (generating the word {\em tokens} in each sentence) with an RNN-based spelling model (generating the letters in each word {\em type}).
These two RNNs respectively capture sentence structure and word structure, and are kept separate as in linguistics.  By invoking the second RNN to generate spellings for novel words in context, we obtain an open-vocabulary language model.  For known words, embeddings are naturally inferred by combining evidence from type spelling and token context.
Comparing to baselines (including a novel strong baseline), we beat previous work and establish state-of-the-art results on multiple datasets.
\end{abstract}

\begin{figure*}[ht]
    \centering
    \newcommand{\lstmnodetext}{\scalebox{.3}{\scalebox{.9}{\begin{varwidth}{5em}\centering\textsf{\textcolor{gray}{\rule{0pt}{1em}RNN\\[0em]cell\rule[-.2em]{0pt}{1em}}}\end{varwidth}}}}
    \newcommand\lstmnode[1]{\node[draw,draw=gray,fill=gray!50,inner sep=0.1em](#1){\lstmnodetext};}
    \newcommand{\lightbulb}[1][black]{\raisebox{-.2em}{\begin{tikzpicture}\draw[draw=none,fill=#1,rotate=180,scale=0.023,rounded corners=0] svg "M242.606,60.651c-66.989,0-121.302,54.315-121.302,121.304c0,60.651,60.651,122.25,60.651,181.951h121.3 c0-61.596,60.653-121.3,60.653-181.951C363.908,114.966,309.598,60.651,242.606,60.651z M306.662,258.625 c-12.083,24.021-24.433,48.664-30.266,74.959h-67.729c-5.834-25.497-17.917-49.752-29.734-73.446 c-14.039-28.222-27.305-54.875-27.305-78.183c0-50.167,40.812-90.978,90.978-90.978c50.166,0,90.979,40.811,90.979,90.978 C333.586,204.997,320.49,231.058,306.662,258.625z M303.255,439.727c0,8.378-6.776,15.159-15.158,15.159h-2.786 c-6.245,17.65-22.925,30.326-42.704,30.326c-19.784,0-36.457-12.676-42.706-30.326h-2.784c-8.38,0-15.161-6.781-15.161-15.159 c0-8.382,6.781-15.168,15.161-15.168h90.98C296.479,424.559,303.255,431.345,303.255,439.727z M303.255,394.237 c0,8.383-6.776,15.159-15.158,15.159h-90.98c-8.38,0-15.161-6.776-15.161-15.159c0-8.382,6.781-15.163,15.161-15.163h90.98 C296.479,379.074,303.255,385.855,303.255,394.237z M92.606,77.831l27.305,15.781c-5.922,8.205-11.045,16.911-15.25,26.21 l-27.219-15.698L92.606,77.831z M242.606,30.327c-5.211,0-10.098,1.008-15.164,1.54V0h30.327v31.867 C252.706,31.334,247.819,30.327,242.606,30.327z M154.263,59.26l-15.782-27.305l26.297-15.166l15.697,27.216 C171.174,48.213,162.47,53.337,154.263,59.26z M380.584,119.822c-4.21-9.299-9.357-18.005-15.28-26.206l27.303-15.786 l15.163,26.299L380.584,119.822z M330.945,59.232c-8.173-5.895-16.909-11.019-26.174-15.226l15.695-27.216l26.238,15.166 L330.945,59.232z M92.576,197.119H60.651v-30.324h31.867c-0.532,5.063-1.54,9.948-1.54,15.161 C90.978,187.108,91.986,192.085,92.576,197.119z M424.562,166.794v30.324h-31.923c0.588-5.034,1.597-10.011,1.597-15.164 c0-5.213-1.009-10.097-1.54-15.161H424.562z M380.229,243.911l27.54,15.873l-15.163,26.271l-24.819-14.332 C372.258,262.565,376.406,253.329,380.229,243.911z M104.867,243.967c3.762,9.422,7.909,18.719,12.349,27.9l-24.609,14.188 l-15.164-26.271L104.867,243.967z";\end{tikzpicture}}}
    \begin{adjustbox}{width=\textwidth}
    \begin{tikzpicture}[node distance = 2em]
        \node[draw=gray, rectangle] (N01) {\textcolor{gray}{$\mathcal{N}(0,I)$}};
        \node[draw, ellipse, inner sep=.1em, right=of N01, fill=black] (eUNK) {\textcolor{white}{$e(\textsc{unk})$}};
        \node[draw, ellipse, inner sep=.1em, below=of eUNK, fill=green!50, xshift=-3em, yshift=1em] (e1) {$e(\circnum{1})$};
        \node[draw, ellipse, inner sep=.1em, below=of e1, fill=violet!50, yshift=-2em] (e2) {$e(\circnum{2})$};
        \node[draw, ellipse, inner sep=.1em, below=of e2, fill=red!50, yshift=-2em] (e3) {$e(\circnum{3})$};

        \draw[gray, -latex] (N01) -- (eUNK);
        \draw[gray, -latex] (N01) |- (e1);
        \draw[gray, -latex] (N01) |- (e2);
        \draw[gray, -latex] (N01) |- (e3);

        \matrix (lexthe) [matrix of nodes, below right=of e1, xshift=-.25em, yshift=1.8em, column sep=0em, ampersand replacement=\&] {
            \lstmnode{lexthe-1-1} \& \lstmnode{lexthe-1-2} \& \lstmnode{lexthe-1-3} \\
            \charword{t} \& \charword{h} \& \charword{e} \\
        };
        \begin{pgfonlayer}{background}
            \node[rounded corners=.1em, fill=gray!30, inner sep=.2em, fit={(lexthe-1-1) (lexthe-1-3)}] {};
        \end{pgfonlayer}
        \draw[->, rounded corners=.2em, thick] (e1.east) -| (lexthe-1-1);
        \draw[->, rounded corners=.2em, thick] (e1.east) -| (lexthe-1-2);
        \draw[->, rounded corners=.2em, thick] (e1.east) -| (lexthe-1-3);
        \draw[gray, ->] ($(lexthe-1-1.west) - (.6em,0)$) -- (lexthe-1-1.west);
        \draw[gray, ->] (lexthe-1-1.east) -- (lexthe-1-2.west);
        \draw[gray, ->] (lexthe-1-2.east) -- (lexthe-1-3.west);
        \draw[gray, ->] (lexthe-1-3.east) -- ($(lexthe-1-3.east) + (.6em,0)$);
        \draw [thick, decoration={brace, mirror, raise=0.5cm}, decorate] (lexthe-2-1.north) -- node[anchor=north, draw, ellipse, inner sep=.1em, fill=green!50, yshift=-1.6em] (s1) {$\sigma(\circnum{1})$} (lexthe-2-3.north);

        \matrix (lexcat) [matrix of nodes, below right=of e2, xshift=-.25em, yshift=1.8em, column sep=0em, ampersand replacement=\&] {
            \lstmnode{lexcat-1-1} \& \lstmnode{lexcat-1-2} \& \lstmnode{lexcat-1-3} \\
            \charword{c} \& \charword{a} \& \charword{t} \\
        };
        \begin{pgfonlayer}{background}
            \node[rounded corners=.1em, fill=gray!30, inner sep=.2em, fit={(lexcat-1-1) (lexcat-1-3)}] {};
        \end{pgfonlayer}
        \draw[->, rounded corners=.2em, thick] (e2.east) -| (lexcat-1-1);
        \draw[->, rounded corners=.2em, thick] (e2.east) -| (lexcat-1-2);
        \draw[->, rounded corners=.2em, thick] (e2.east) -| (lexcat-1-3);
        \draw[gray, ->] ($(lexcat-1-1.west) - (.6em,0)$) -- (lexcat-1-1.west);
        \draw[gray, ->] (lexcat-1-1.east) -- (lexcat-1-2.west);
        \draw[gray, ->] (lexcat-1-2.east) -- (lexcat-1-3.west);
        \draw[gray, ->] (lexcat-1-3.east) -- ($(lexcat-1-3.east) + (.6em,0)$);
        \draw [thick, decoration={brace, mirror, raise=0.5cm}, decorate] (lexcat-2-1.north) -- node[anchor=north, draw, ellipse, inner sep=.1em, fill=violet!50, yshift=-1.6em] (s2) {$\sigma(\circnum{2})$} (lexcat-2-3.north);

        \matrix (lexchased) [matrix of nodes, below right=of e3, xshift=-.25em, yshift=1.8em, column sep=0em, ampersand replacement=\&] {
            \lstmnode{lexchased-1-1} \& \lstmnode{lexchased-1-2} \& \lstmnode{lexchased-1-3} \& \lstmnode{lexchased-1-4} \& \lstmnode{lexchased-1-5} \& \lstmnode{lexchased-1-6} \\
            \charword{c} \& \charword{h} \& \charword{a} \& \charword{s} \& \charword{e} \& \charword{d} \\
        };
        \begin{pgfonlayer}{background}
            \node[rounded corners=.1em, fill=gray!30, inner sep=.2em, fit={(lexchased-1-1) (lexchased-1-6)}] {};
        \end{pgfonlayer}
        \draw[->, rounded corners=.2em, thick] (e3.east) -| (lexchased-1-1);
        \draw[->, rounded corners=.2em, thick] (e3.east) -| (lexchased-1-2);
        \draw[->, rounded corners=.2em, thick] (e3.east) -| (lexchased-1-3);
        \draw[->, rounded corners=.2em, thick] (e3.east) -| (lexchased-1-4);
        \draw[->, rounded corners=.2em, thick] (e3.east) -| (lexchased-1-5);
        \draw[->, rounded corners=.2em, thick] (e3.east) -| (lexchased-1-6);
        \draw[gray, ->] ($(lexchased-1-1.west) - (.6em,0)$) -- (lexchased-1-1.west);
        \draw[gray, ->] (lexchased-1-1.east) -- (lexchased-1-2.west);
        \draw[gray, ->] (lexchased-1-2.east) -- (lexchased-1-3.west);
        \draw[gray, ->] (lexchased-1-3.east) -- (lexchased-1-4.west);
        \draw[gray, ->] (lexchased-1-4.east) -- (lexchased-1-5.west);
        \draw[gray, ->] (lexchased-1-5.east) -- (lexchased-1-6.west);
        \draw[gray, ->] (lexchased-1-6.east) -- ($(lexchased-1-6.east) + (.6em,0)$);
        \draw [thick, decoration={brace, mirror, raise=0.5cm}, decorate] (lexchased-2-1.north) -- node[anchor=north, draw, ellipse, inner sep=.1em, fill=red!50, yshift=-1.6em] (s3) {$\sigma(\circnum{3})$} (lexchased-2-6.north);


        \begin{pgfonlayer}{backgroundoverlay}
            \draw[dotted, gray!50, thick, fill=gray, fill opacity=.05, rounded corners = 1em]
                ($(e3) + (-2.5em,-1.5em)$) --
                ($(e1) + (-2.5em,  .5em)$) --
                ($(eUNK) + (-2em, 1.5em)$) --
                ($(eUNK) + ( 3em, 1.5em)$)--
                ($(eUNK) + ( 3em,-.75em)$) --
                ($(e1) + ( 2em,   0em)$) --
                ($(e3) + ( 2em,-1.5em)$) --
                cycle;
            \node[below right=of eUNK, xshift=7em, yshift=-5.7em](embeddings){};
            \node[below=of embeddings, yshift=2.35em]{\textcolor{gray!50}{\scriptsize look up embeddings}};
            \draw[dotted, gray!50, thick, fill=gray, fill opacity=.05, rounded corners = 1em]
                ($(s1) + (-2.5em, 1.5em)$) --
                ($(s2) + (-2.5em,   0em)$) --
                ($(s3) + (  -2em,-1.5em)$) --
                ($(s3) + (   3em,-1.5em)$) --
                ($(s2) + ( 2.5em,   0em)$) --
                ($(s1) + ( 2.5em, 1.5em)$) --
                cycle;
            \node[below=of embeddings, yshift=-.25em]{\textcolor{gray!50}{\scriptsize look up spellings}};
        \end{pgfonlayer}

        \node[draw, circle, right=of eUNK, xshift = 6em, yshift=-5em](h1){$\vec{h}_1$};
        \node[draw, ellipse, dashed, fill=green!25, inner sep=.2em, below right=of h1, xshift=.75em, yshift=-2em] (tok1) {\begin{varwidth}{\textwidth}$w_1$\\\tiny\smash{\raisebox{-.2em}{$= \circnum{1}$}}\end{varwidth}};
        \draw[-latex, decorate, decoration={snake,segment length=.5em,amplitude=.1em,post=lineto,post length=.5em}] (h1) to node[above,rotate=-54,xshift=-.5em](softmax1){\scriptsize softmax} (tok1);

        \begin{pgfonlayer}{backgroundoverlay}
            \draw[dotted, -latex, gray!50, thick, rounded corners=1em] ($(eUNK) + (3em,.4em)$) to[out=340, in=180] (embeddings.center) -- (h1 |- embeddings.center) -- (softmax1.south);
        \end{pgfonlayer}

        \node[draw, circle, right=of h1, xshift = 3em](h2){$\vec{h}_2$};
        \draw[-latex] (h1) to node[below,xshift=1.5em]{\scriptsize RNN} (h2);
        \node[draw, ellipse, dashed, fill=violet!25, inner sep=.2em, below right=of h2, xshift=.75em, yshift=-2em] (tok2) {\begin{varwidth}{\textwidth}$w_2$\\\tiny\smash{\raisebox{-.2em}{$= \circnum{2}$}}\end{varwidth}};
        \draw[-latex, decorate, decoration={snake,segment length=.5em,amplitude=.1em,post=lineto,post length=.5em}] (h2) to node[above,rotate=-54,xshift=-.5em](softmax2){\scriptsize softmax} (tok2);
        \begin{pgfonlayer}{backgroundoverlay}
            \draw[dotted, -latex, gray!50, thick, rounded corners=1em] ($(eUNK) + (3em,.4em)$) to[out=340, in=180] (embeddings.center) -- (h2 |- embeddings.center) -- (softmax2.south);
            \draw[dotted, gray!50, thick, rounded corners=1em] ($(eUNK) + (3em,.4em)$) to[out=340, in=180] (embeddings.center) -- (tok1.east |- embeddings.center) -- (h2.south west);
        \end{pgfonlayer}
        \draw[-latex] (tok1) to (h2.south west);

        \node[draw, circle, right=of h2, xshift = 3em](h3){$\vec{h}_3$};
        \draw[-latex] (h2) to node[below,xshift=1.5em]{\scriptsize RNN} (h3);
        \node[ellipse, fill=black, inner sep=.2em, below right=of h3, xshift=.75em, yshift=-2em] (tok3) {\begin{varwidth}{\textwidth}\centering\textcolor{white}{$w_3$}\\[-.2em]\tiny\smash{\raisebox{-.2em}{\textcolor{white}{$=$ \textsc{unk}}}}\end{varwidth}};
        \draw[-latex, decorate, decoration={snake,segment length=.5em,amplitude=.1em,post=lineto,post length=.5em}] (h3) to node[above,rotate=-54,xshift=-.5em](softmax3){\scriptsize softmax} (tok3);
        \begin{pgfonlayer}{backgroundoverlay}
            \draw[dotted, -latex, gray!50, thick, rounded corners=1em] ($(eUNK) + (3em,.4em)$) to[out=340, in=180] (embeddings.center) -- (h3 |- embeddings.center) -- (softmax3.south);
            \draw[dotted, gray!50, thick, rounded corners=1em] ($(eUNK) + (3em,.4em)$) to[out=340, in=180] (embeddings.center) -- (tok2.east |- embeddings.center) -- (h3.south west);
        \end{pgfonlayer}
        \draw[-latex] (tok2) to (h3.south west);

        \node[draw, circle, right=of h3, xshift = 3em](h4){$\vec{h}_4$};
        \draw[-latex] (h3) to node[below,xshift=1.5em]{\scriptsize RNN} (h4);
        \node[draw, ellipse, dashed, fill=green!25, inner sep=.2em, below right=of h4, xshift=.75em, yshift=-2em] (tok4) {\begin{varwidth}{\textwidth}$w_4$\\\tiny\smash{\raisebox{-.2em}{$= \circnum{1}$}}\end{varwidth}};
        \draw[-latex, decorate, decoration={snake,segment length=.5em,amplitude=.1em,post=lineto,post length=.5em}] (h4) to node[above,rotate=-54,xshift=-.5em](softmax4){\scriptsize softmax} (tok4);
        \begin{pgfonlayer}{backgroundoverlay}
            \draw[dotted, -latex, gray!50, thick, rounded corners=1em] ($(eUNK) + (3em,.4em)$) to[out=340, in=180] (embeddings.center) -- (h4 |- embeddings.center) -- (softmax4.south);
            \draw[dotted, gray!50, thick, rounded corners=1em] ($(eUNK) + (3em,.4em)$) to[out=340, in=180] (embeddings.center) -- (tok3 |- embeddings.center) -- (h4.south west);
        \end{pgfonlayer}
        \draw[-latex] (tok3) to (h4.south west);

        \node[draw, circle, right=of h4, xshift = 3em](h5){$\vec{h}_5$};
        \draw[-latex] (h4) to node[below,xshift=1.5em]{\scriptsize RNN} (h5);
        \node[circle, inner sep=.05em, below right=of h5, xshift=.75em, yshift=-2em] (tok5) {...};
        \draw[-latex, decorate, decoration={snake,segment length=.5em,amplitude=.1em,post=lineto,post length=.5em}] (h5) to node[above,rotate=-54,xshift=-.5em](softmax5){\scriptsize softmax} (tok5);
        \begin{pgfonlayer}{backgroundoverlay}
            \draw[dotted, -latex, gray!50, thick, rounded corners=1em] ($(eUNK) + (3em,.4em)$) to[out=340, in=180] (embeddings.center) -- (h5 |- embeddings.center) -- (softmax5.south);
            \draw[dotted, gray!50, thick, rounded corners=1em] ($(eUNK) + (3em,.4em)$) to[out=340, in=180] (embeddings.center) -- (tok4.east |- embeddings.center) -- (h5.south west);
        \end{pgfonlayer}
        \draw[-latex] (tok4) to (h5.south west);

        \matrix (textliked) [matrix of nodes, below=of tok3, yshift=-0em, column sep=0em, ampersand replacement=\&] {
            \lstmnode{textliked-1-1} \& \lstmnode{textliked-1-2} \& \lstmnode{textliked-1-3} \& \lstmnode{textliked-1-4} \& \lstmnode{textliked-1-5} \\[.75em]
            \charword{c} \& \charword{a} \& \charword{g} \& \charword{e} \& \charword{d} \\
        };
        \begin{pgfonlayer}{background}
            \node[rounded corners=.1em, fill=gray!30, inner sep=.2em, fit={(textliked-1-1) (textliked-1-5)}] {};
        \end{pgfonlayer}
        \node(h3break) at ($(h3) + (0em,-7.75em)$){};
        \draw[->, rounded corners=.2em] (h3) -- (h3break.center) -| (textliked-1-1);
        \draw[->, rounded corners=.2em] (h3) -- (h3break.center) -| (textliked-1-2);
        \draw[->, rounded corners=.2em] (h3) -- (h3break.center) -| (textliked-1-3);
        \draw[->, rounded corners=.2em] (h3) -- (h3break.center) -| (textliked-1-4);
        \draw[->, rounded corners=.2em] (h3) -- (h3break.center) -| (textliked-1-5);
        \draw[gray, ->] ($(textliked-1-1.west) - (.6em,0)$) -- (textliked-1-1.west);
        \draw[gray, ->] (textliked-1-1.east) -- (textliked-1-2.west);
        \draw[gray, ->] (textliked-1-2.east) -- (textliked-1-3.west);
        \draw[gray, ->] (textliked-1-3.east) -- (textliked-1-4.west);
        \draw[gray, ->] (textliked-1-4.east) -- (textliked-1-5.west);
        \draw[gray, ->] (textliked-1-5.east) -- ($(textliked-1-5.east) + (.6em,0)$);

        \node[minimum height=2em, minimum width=3em](textcat) at (textliked-2-5 -| tok2) {\charword{cat}};
        \node[minimum height=2em, minimum width=3em](textthe1) at (textliked-2-5 -| tok1) {\charword{the}};
        \node[minimum height=2em, minimum width=3em](textthe2) at (textliked-2-5 -| tok4) {\charword{the}};
        \node[minimum height=2em, minimum width=3em](textrest) at (textliked-2-5 -| tok5) {...};

        \begin{pgfonlayer}{backgroundoverlay}
            \draw[yellow!80!brown, thick, fill=yellow!80!brown, opacity=.2, fill opacity=.1]
                (textthe1.north west) --
                (textrest.north east) --
                ($(textrest.south east) + (0em,.2em)$) --
                ($(textrest.south east) + (-.2em,0em)$) --
                (textthe1.south west) --
                cycle;
        \end{pgfonlayer}

        \begin{pgfonlayer}{backgroundoverlay}
            \draw[dotted, gray!50, thick, rounded corners=2em] ($(s2) + (2.5em,0em)$) -- ($(s2) + (5em,0.5em)$) -| (textthe1);
            \draw[dotted, gray!50, thick, rounded corners=2em] ($(s2) + (2.5em,0em)$) -- ($(s2) + (5em,0.5em)$) -| (textcat);
            \draw[dotted, gray!50, thick, rounded corners=2em] ($(s2) + (2.5em,0em)$) -- ($(s2) + (5em,0.5em)$) -| (textthe2);
        \end{pgfonlayer}

        \draw[-latex] (tok1) to (textthe1);
        \draw[-latex] (tok2) to (textcat);
        \draw[-latex] (tok4) to (textthe2);

        \node[minimum height=2em, minimum width=3em, below=of textthe1, yshift=2em] {$\sigma(w_1)$};
        \node[minimum height=2em, minimum width=3em, below=of textcat, yshift=2em] {$\sigma(w_2)$};
        \node[minimum height=2em, minimum width=3em, below=of textliked-2-3, yshift=1.7em] {$\spelling \sim p_\mathrm{spell}(\cdot \mid \vec{h}_3)$};
        \node[minimum height=2em, minimum width=3em, below=of textthe2, yshift=2em] {$\sigma(w_4)$};

        \node[left=of N01, xshift=.5em] {};
        \node[right=of h4, xshift=11em] {};

        \node[right=of eUNK, xshift=21em, yshift=-1em, draw=none, fill=gray!5, inner sep=.5em, rounded corners=.5em, scale=.9] {\scalebox{1.3}{\lightbulb[gray]{}}\;\, \parbox{15em}{\centering\em\color{gray} {\Large usage $\perp$ spelling $\mid$ embedding} \\[-.2em] {\scriptsize\color{gray!50} (conditional independence property)}} \;\,\scalebox{1.3}{\lightbulb[gray]{}}};
        \node[right=of eUNK, xshift=5.5em, yshift=-1em, draw=none, fill=gray!5, inner sep=.5em, rounded corners=.5em, scale=.9] {\scalebox{1.3}{\lightbulb[gray]{}}\;\, \parbox{10em}{\centering\em\color{gray} \footnotesize \textbf{regularize}, don't \textbf{construct}\\embeddings using spellings} \;\,\scalebox{1.3}{\lightbulb[gray]{}}};
    \end{tikzpicture}
    \end{adjustbox}
    \caption{
    A lexeme's embedding is optimized to be predictive of the lexeme's spelling.
    That spelling is predicted only once (left); every corpus token of that lexeme type (right) simply ``summons,'' or copies, the type's spelling.
    For the novel word $w_3$ (\cref{sec:spelling-unks}), \charword{caged} is preferred over something unpronounceable like \charword{xsmfk}, but also over the ungrammatical \charword{furry}, because the hidden state $\vec{h_3}$ prefers a verb and the spelling model can generalize from the verb \charword{chased} ending in \charword{-ed} to \charword{caged}.}
    \label{fig:model}
\end{figure*}
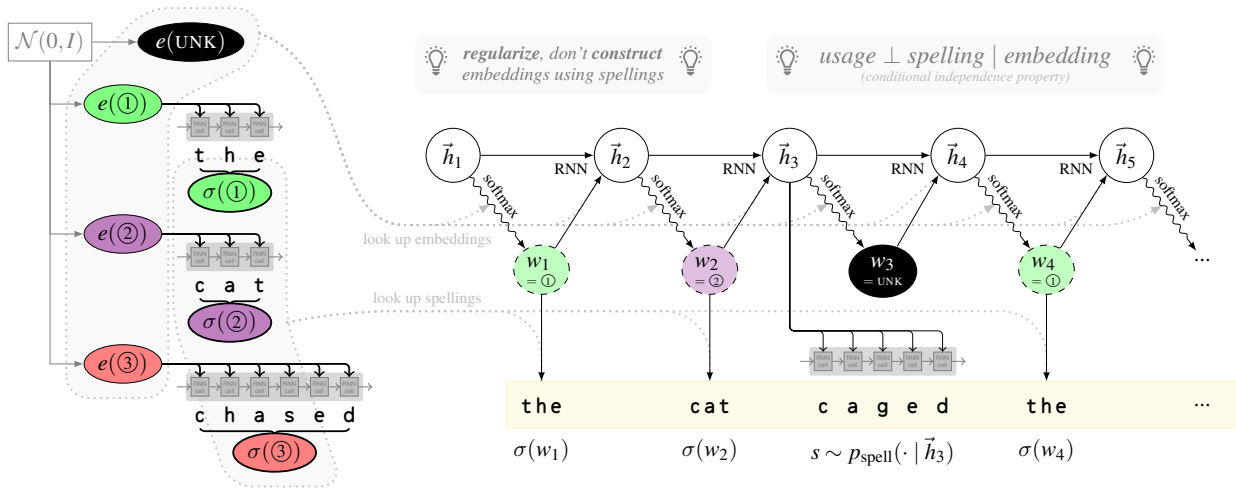

\section{Introduction}\label{sec:introduction}

In this paper, we propose a neural language model that incorporates a generative model of word spelling.
That is, we aim to explain the training corpus as resulting from a process that first generated a lexicon of word types---the language's vocabulary---and then generated the corpus of tokens by ``summoning'' those types.

Each entry in the lexicon specifies both a syntactic/semantic embedding vector and an orthographic spelling.  Our model captures the correlation between these, so it can extrapolate to predict the spellings of \emph{unknown} words in any syntactic/semantic context.  In this sample from our trained English model, the words in \wughighlight{\small this font} were unobserved in training data, yet have contextually appropriate spellings:

\begin{quote}
    \sffamily\color{gray!30!black}
    Following the death of Edward McCartney in \wughighlight{1060} , the new definition was transferred to the \wughighlight{WDIC} of \wughighlight{Fullett} .
\end{quote}

While the fully generative approach is shared by previous Bayesian models of language (e.g., \citet{goldwater-et-al-2006}), even those that model characters and words at different levels \citep{MocYamUed09Bayesian,GolGriJoh11Producing} have no embeddings and hence no way to relate
spelling to usage.  They also have an impoverished model of sequential structure (essentially,
$n$-gram models with backoff).  We instead employ \emph{recurrent neural networks} 
to model both the sequence of words in a sentence and the sequence of characters in a word type, where
the latter sequence is conditioned on the word's embedding.
The resulting model achieves state-of-the-art on multiple datasets. It is well-founded in linguistics and Bayesian modeling, but we can easily use it in the traditional non-Bayesian language modeling setting by performing MAP estimation.

We begin by concisely stating a first, closed-vocabulary, version of our model in \cref{sec:embs-and-spellings}, before explaining our motivations from various perspectives in \cref{sec:words-characters-types-tokens}.
Then \Cref{sec:spelling-unks} motivates and describes a simple way to extend the model to the open-vocabulary setting.
\cref{sec:experiments} contains quantitative and qualitative experiments on multiple language modeling datasets, with implementation details provided in supplementary material.
Finally, we clarify the relation of this model to previous work in \cref{sec:previous-work} and summarize our contribution in \cref{sec:conclusion-future}.

\section{A joint model of lexicon and text}\label{sec:embs-and-spellings}

\subsection{Lexemes have embeddings and spellings}

We will model text that has already been tokenized, i.e., it is presented as a sequence of word tokens $w_1,w_2,\ldots$.

We assume that a language's word types, which we henceforth call \emph{lexemes} to avoid confusion, are discrete elements $w$ of the {\em vocabulary} $\mathcal{V} = \{ \circnum{1}, \circnum{2}, \ldots \circnum{$v$} \}$.
In our model, the observable behavior of the lexeme $w$ is determined by two properties: a latent real-valued \emph{embedding} $\ofw{e} \in \mathbb{R}^d$, which governs where $w$ tends to appear, and $w$'s spelling $\ofw{\sigma} \in \Sigma^*$ (for some alphabet of characters  $\Sigma$),
which governs how it looks orthographically when it does appear.

We will use $e$ and $\sigma$ to refer to the functions that map each lexeme $w$ to its embedding and spelling.  Thus the lexicon is specified by $(e, \sigma)$.  Our model\footnote{Before extension to the open-vocabulary case, found in \cref{eqn:ovloss}.} (given fixed $v$ and $n$) specifies a joint distribution over the lexicon and corpus:
\begin{gather}\label{eq:generative}
    p(\theta, e, \sigma, w_1, \ldots, w_n) \;\; =\;\; p(\theta) \;\cdot\\
    \underbrace{\prod_{w \in \mathcal{V}} \Big[
        \hspace{-.4em}
        \underbrace{p(\!\ofw{e}\!)}_{\substack{\text{prior on}\\\text{embeddings}}}
        \hspace{-.4em} \cdot
        \underbrace{p_\mathrm{spell}(\ofw{\sigma} \!\mid\! \ofw{e})}_{\substack{\text{spelling model}\\\text{for all types}}}
    \Big]}_{\text{lexicon generation}}
    \hspace{-.1em} \cdot \hspace{-.1em}
    \underbrace{\prod_{i=1}^n p_\mathrm{LM}(w_i \!\mid\! \vec{w}_{<i}, e)}_{\substack{\text{\phantom{g}lexeme-level\phantom{g}}\\\text{recurrent language model}\\\text{for all tokens}}}\nonumber
\end{gather}
where $p_{\text{LM}}$ (\cref{sec:lexeme-level-rnn}) and $p_{\text{spell}}$ (\cref{sec:speller-model}) are RNN sequence models that are parameterized by $\theta$ (the dependence is omitted for space reasons), and
$w_1, \ldots, w_n$ is a sequence of word tokens.

Let us unpack this formula.
The generative story for the observed training corpus has two steps:
\begin{description}
\item[Generate the structure of the language.] First draw RNN parameters $\theta$ (from a spherical Gaussian). Then draw an embedding $\ofw{e}$ for each lexeme $w$ (from another spherical Gaussian).\footnote{We will later find the MAP estimates of $\theta$ and $\ofw{e}$, so the Gaussian priors correspond to $L_2$ regularization of these estimates.}  Finally, sample a spelling $\ofw{\sigma}$ for each lexeme $w$ from the $p_{\text{spell}}$ model, conditioned on $\ofw{e}$ (and $\theta$).
\item[Generate the corpus.]  In the final term of \eqref{eq:generative}, generate a sequence of lexemes $w_1, \ldots, w_n$ from the $p_{\text{LM}}$ model (using $\theta$).  Applying $\sigma$ yields the actually observed training corpus $\sigma(w_1), \ldots, \sigma(w_n)$, a sequence of spelled words.
\end{description}

In the present paper we make the common simplifying assumption that the training corpus has no polysemy, so that two word tokens with the same spelling always correspond to the same lexeme.  We thus assign distinct lexeme numbers $\circnum{1}, \circnum{2}, \ldots, \circnum{v}$ to the different spelling types in the corpus (the specific assignment does not matter).  Thus, we have observed the spellings $\sigma(\circnum{1}), \sigma(\circnum{2}), \ldots, \sigma(\circnum{v})$ of these $v$ assumed lexemes.  We have also observed the actual token sequence $w_1,\ldots,w_n$ where each $w_i$ is a lexeme number.

Given these observations, we train $\theta$ and $e$ jointly by MAP estimation: in other words, we choose them to (locally) maximize \eqref{eq:generative}.\footnote{The non-Bayesian view on this is that we maximize the \emph{regularized likelihood} of the model given the observations.}  This is straightforward using backpropagation and gradient descent (see \cref{sec:joint-model}\footnote{All appendices (supp.\@ material) can be found after the references on page \pageref{sec:joint-model}.} on training and \cref{sec:hyperparameters} for implementation and hyperparameter details).

\subsection{Modeling word sequences with $p_\mathrm{LM}$}\label{sec:lexeme-level-rnn}

The final term of \eqref{eq:generative} is simply a neural language model that uses the embeddings $e$.  It should be stressed that \emph{any} such neural language model can be used here. We will use the commonly used AWD-LSTM built by \citet{MerKesSoc17Regularizing} with their best reported hyperparameters.

\subsection{Modeling letter sequences with $p_{\text{spell}}$}\label{sec:speller-model}

Our model also has to predict the spelling of every lexeme.  We model $p_{\text{spell}}(\ofw{\sigma} \mid \ofw{e})$ with a vanilla LSTM language model \cite{SunSchNey12LSTM}, this time over characters, using special characters \textsc{bow} (beginning of word) and \textsc{eow} (end of word) to begin and end the sequence.

Our intuition is that in most languages, the spelling of a word tends to weakly reflect categorical properties that are hopefully captured in the embedding.  For example, proper names may have a certain form, content words may have to contain at least two syllables \cite{mccarthy-prince-1999}, and past-tense verbs may tend to end with a certain suffix.  This is why $p_{\text{spell}}(\ofw{\sigma} \mid \ofw{e})$ is conditioned on the lexeme's embedding $\ofw{e}$.  We accomplish this by feeding $\ofw{e}$ into $\mathrm{LSTM}_\mathrm{spell}$
as additional input at every step, alongside the ordinary inputs (the previous hidden state $\vec{h_{t-1}}$ and an low-dimensional embedding $\vec{c_{t-1}} \in \mathbb{R}^{d'}$ of the previous character):\\[-.5em]
\begin{equation}
    \vec{h_t} = \mathrm{LSTM}_\mathrm{spell}\big( \; \vec{h_{t-1}}, \; [\,\vec{c_t}\,;\, \ofw{e}\,] \; \big)
\end{equation}

As the spelling model tends to overfit to training lexemes (instead of modeling the language's phonotactics, morphology, and other conventions),
we project the embeddings $\ofw{e}$ into a \emph{lower-dimensional space} to combat this overfitting.
We do so by regularizing the four input weight matrices\footnote{$W_{i\,i}$, $W_{i\!f}$, $W_{ig}$, and $W_{io}$ in PyTorch's \texttt{LSTMCell} documentation; regularization only applies to the part that is multiplied with $\ofw{e}$.}
of $\mathrm{LSTM}_\mathrm{spell}$ with the \emph{nuclear norm} (the sum of each matrix's singular values; details on it in \cref{sec:nuclearnorm}), resulting in a low-rank projection.
The nuclear norm (times a positive hyperparameter) is added to the objective, namely the negative log of \cref{eq:generative}, as part of the definition of the $-\!\log p(\theta)$ term.
This regularizer indeed helps on development data, and it outperformed $L_2$ regularization in our pilot experiments.

\section{Words, characters, types, and tokens}\label{sec:words-characters-types-tokens}

We now take a step back and discuss our modeling principles.  Our model structure above aims to incorporate two perspectives that have been neglected in neural language modeling:

\subsection{A linguistic perspective}\label{sec:duality-of-patterning}

\citet{Hoc60Origin} regarded \emph{duality of patterning} as a fundamental property of human language:
the \emph{form} of a word is logically separate from its \emph{usage}.
For example, while \charword{children} may be an unusual spelling for a plural noun in English, it is listed as one in the lexicon, and that grants it all the same privileges as any other plural noun.
The syntactic and semantic processes that combine words are blind to its unusual spelling.
In our two-level architecture, this ``separation of concerns'' holds between $p_{\text{spell}}$, which governs word \emph{form}, and $p_{\text{LM}}$, which governs word \emph{usage}.
A word's distribution of contexts is conditionally independent of its spelling, given its embedding,
because $p_{\text{LM}}$ does not consult spellings but only embeddings.
Prior work does {\em not} do this---character-based language models blur the two levels into one.

Half a century earlier, \citet{Sau16Course} discussed the {\em arbitrariness of the sign}.  In our model, $p_{\text{spell}}$ has support on all of $\Sigma^*$, so any embedding can in principle be paired in the lexicon with any spelling---even if some pairings may be more likely {\em a priori} than others, perhaps because they are more pronounceable or the result of a morphological transformation.
In contrast with most prior work that compositionally derives a word's embedding from its spelling, our model only {\em prefers} a word's embedding to correlate with its spelling, in order to raise the factor $p_{\text{spell}}(\ofw{\sigma} \mid \ofw{e})$.  This preference is merely a regularizing effect that may be overcome by the need to keep the $p_{\text{LM}}$ factors large, particularly for a frequent word that appears in many $p_{\text{LM}}$ factors and is thus allowed its own idiosyncratic embedding.

In short, our spelling model is \emph{not} supposed to be able to perfectly predict the spelling.  However, it can statistically capture phonotactics, regular and semi-regular inflection, etc.

\subsection{A modeling perspective}\label{sec:typetoken}

The distinction between word types (i.e., entries in a vocabulary) and tokens (i.e., individual occurrences of these word types in text) is also motivated by a generative (e.g., Bayesian) treatment of language: a lexeme's spelling is \emph{reused} over all its tokens, not generated from scratch for each token.

This means that the term $p_{\text{spell}}(\ofw{\sigma} \mid \ofw{e})$ appears only once in \eqref{eq:generative} for each word type $w$.
Thus, the training of the spelling model is not \emph{overly} influenced by frequent (and atypical) words like \charword{the} and \charword{a},%
\footnote{%
The striking difference between types and tokens is perhaps most visible with \charword{th}.
It is the most common character bigram in words of the Penn Treebank as preprocessed by \citet{MikKarBur10Recurrent} when counting word \emph{tokens}, but only appears in \nth{156} place when counting word \emph{types}.
Looking at trigrams (with spaces) produces an even starker picture: \charword{\textvisiblespace{}th}, \charword{the}, \charword{he\textvisiblespace{}} are respectively the \nth{1}, \nth{2}, and \nth{3} most common trigrams when looking at tokens, but only the \nth{292}, \nth{550}, and \nth{812} (out of 5261) when considering types.
}
but just as much as by rare words like \charword{deforestation}.
As a result, $p_{\text{spell}}$ learns how typical word {\em types}---not typical {\em tokens}---are spelled.  This is useful in predicting how {\em other types} will be spelled, which helps us regularize the embeddings of rare word types and predict the spellings of novel word types (\cref{sec:spelling-unks} below).%
\footnote{\citet{BaaSpr96Estimating} argue for using only the \emph{hapax legomena} (words that only appear once) to predict the behavior of rare and unknown words.  The Bayesian approach \cite{MacPet95Hierarchical,Teh06Hierarchical} is a compromise: frequent word types are also used, but they have no more influence than infrequent ones.}

\section{Open vocabulary by ``spelling'' \textsc{unk}}\label{sec:spelling-unks}

Our spelling model not only helps us regularize the embeddings of rare words, but also allows us to handle \emph{unknown words}, a long-standing concern in NLP tasks.%
\footnote{Often 5--10\% of held-out word tokens
in language modeling datasets were never seen in training data.  Rates of 20--30\% or more can be encountered if the model was trained on out-of-domain data.}
How?

As a language usually has a fixed known alphabet (so the held-out data will at least not contain unknown characters), a common approach is to model character sequences instead of word sequences to begin with \citep{SutMarHin11Generating}.  However, such a model does not explicitly represent word units,
does not respect duality of patterning (\cref{sec:duality-of-patterning}), and thus may have a harder time learning syntactic and semantic patterns at the sentence level.  For this reason, several recent approaches have tried to combine character-level modeling with word-level modeling (see \cref{sec:previous-work}).

Our approach differs from this previous work because we have an explicit spelling model to use.  Just as $p_{\text{spell}}$ has an opinion about how to spell rare words, it also has an opinion about how to spell novel words.
This allows the following trick.  We introduce a special lexeme \textsc{unk}, so that the vocabulary is now $\mathcal{V}= \{\textsc{unk}, \circnum{1}, \circnum{2}, \ldots, \circnum{v}\}$ with finite size $v+1$.
We refine our story of how the corpus is generated. First, the model again predicts a complete sequence of lexemes $w_1, \ldots, w_n$.  In most cases, $w_i$ is spelled out deterministically as $\sigma(w_i)$.  However, if $w_i=\textsc{unk}$, then we spell it out by sampling from $p_{\text{spell}}(\cdot \mid \vec{e}_i)$, where $\vec{e}_i$ is an appropriate embedding, explained below.  The downside of this approach is that each \textsc{unk} token samples a fresh spelling, so multiple tokens of an out-of-vocabulary word type are treated as if they were separate lexemes.

Recall that the spelling model generates a spelling \emph{given an embedding}.
So what embedding $\vec{e}_i$ should we use to generate this unknown word?
Imagine the word had been in the vocabulary. Then, if the model had wanted to predict that word, $\vec{e}_i$ would have had to have a high dot product with the hidden state of the lexeme-level RNN at this time step, $\vec{h}$.
So, clearly, the embedding that maximizes the dot product with the hidden state is just the hidden state itself.%
\footnote{At least, this is the best $\vec{e}_i$ for which $|| \vec{e}_i || \leq ||\vec{h}||$ holds.}
It follows that we should sample the generated spelling $\spelling \sim p(\cdot \mid \vec{h})$, using the current hidden state of the lexeme-level RNN.%
\footnote{\label{fn:doublegen}Note that an in-vocabulary token can now be generated in two ways, as the spelling of a known lexeme or of \textsc{unk}.  \Cref{sec:latent-mess} discusses this (largely inconsequential) issue.}

Continuing the generative story, the lexeme-level RNN moves on, but to simplify the inference we feed $e(\textsc{unk})$ into the lexeme-level RNN to generate the next hidden state, rather than feeding in $\vec{h}$ (our guess of $e(\sigma^{-1}(\spelling))$).%
\footnote{This makes the implementation simpler and faster. One could also imagine feeding back, e.g., the final hidden state of the speller.}

Now we can expand the model described in \cref{sec:embs-and-spellings} to deal with sequences containing unknown words.  Our building blocks are two old factors from \cref{eq:generative} and a new one:

\begin{description}
    \item[the lexicon generation]\ $\prod_{w \in \mathcal{V}} \Big[p(e\raisebox{.06em}{\scalebox{.7}{$(w)$}}) \cdot p_\mathrm{spell}(\ofw{\sigma} \!\mid\! \ofw{e})\Big]$\\ 
    predicts the spellings of in-vocabulary lexemes from their embeddings
    \item[the lexeme-level RNN]\ $\prod_{i=1}^n p_\mathrm{LM}(w_i \mid \vec{w}_{<i}, e)$\\
    predicts lexeme $w_i$ from the history $\vec{w}_{<i}$ summarized as $\vec{h}_i$
    \item[the spelling of an \textsc{unk}]\textit{(new!)}\ \ $p_\mathrm{spell}(\spelling \mid \vec{h}')$\\
    predicts the spelling $\spelling$ for an \textsc{unk} lexeme that appears in a context that led the lexeme-level RNN to hidden state $\vec{h}'$
\end{description}

Using these we can again find the MAP estimate of our parameters, i.e., the (regularized) maximum likelihood (ML) solution, using the posterior that is proportional to the new joint (with the change from \cref{eq:generative} in black):
\begin{gather}
    \color{gray} p(\theta, e, \sigma, \spelling_1 \cdots \spelling_n) \;\;=\;\; p(\theta)\; \cdot{}\label{eqn:ovloss}\\
    \color{gray} \prod_{w \in \mathcal{V}} \Big[
        p(\!\ofw{e}\!)
        \cdot
       p_\mathrm{spell}(\ofw{\sigma} \!\mid\! \ofw{e})
    \Big] \cdot \prod_{i=1}^n p_\mathrm{LM}(w_i \mid \vec{w}_{<i}, e) \;\cdot\;\nonumber \\
    \prod_{\hspace*{-1em} i\colon w_i = \textsc{unk} \hspace*{-1em}} p_\mathrm{spell}(\spelling_i \mid \vec{h}_i)\nonumber
\end{gather}
where $\spelling_1, \ldots, \spelling_n$ are the observed spellings that make up the corpus and $w_i = \sigma^{-1}(\spelling_i)$ if defined, i.e.,
if there is a $w \in \mathcal{V}$ with $\sigma(w) = \spelling_i$, and $w_i = \textsc{unk}$ otherwise.

The entire model is depicted in \cref{fig:model}. We train it using SGD, computing the different factors of \cref{eqn:ovloss} in an efficient order (implementation details are presented in \cref{sec:joint-model}).

\section{Experiments}\label{sec:experiments}

We will now describe the experiments we perform to show that our approach works well in practice.%
\footnote{Code at \url{github.com/sjmielke/spell-once}.}
A detailed discussion of all hyperparameters can be found in \cref{sec:hyperparameters}.

\subsection{Datasets}

We evaluate on two open-vocabulary datasets, \emph{WikiText-2} \citep{MerXioBra17Pointer} and the \emph{Multilingual Wikipedia Corpus} \citep{KawDyeBlu17Learning}.%
\footnote{Unlike much previous LM work, we do not evaluate on the Penn Treebank (PTB) dataset as preprocessed by \citet{MikKarBur10Recurrent} as its removal of out-of-vocabulary words makes it fundamentally unfit for open-vocabulary language model evaluation.}
For each corpus, we follow \citeauthor{KawDyeBlu17Learning} and replace characters
that appear fewer than 25 times by a special symbol
$\Diamond$.%
\footnote{This affects fewer than 0.03\% of character tokens of WikiText-2 and thus does not affect results in any meaningful way.}

\subsubsection{WikiText-2}\label{sec:wikitext2}

The WikiText-2 dataset \citep{MerXioBra17Pointer} contains more than 2 million tokens from
the English Wikipedia.
We specifically use the ``raw'' version, which is tokenized but has no
\textsc{unk} symbols (since we need the spellings of \emph{all} words).

The results for WikiText-2 are shown in \cref{tab:results} in the form of bits per character (bpc).  Our full model is denoted \textsc{\bfseries full}. The other rows report on baselines (\cref{sec:baselines}) and ablations (\cref{sec:analysis}), which are explained below.

\subsubsection{Multilingual Wikipedia Corpus}\label{sec:mwc}

The Multilingual Wikipedia Corpus \citep{KawDyeBlu17Learning} contains 360 Wikipedia articles in English, French, Spanish, German, Russian, Czech, and Finnish.
However, we re-tokenize the dataset, not only splitting on spaces (as \citeauthor{KawDyeBlu17Learning} do) but also by splitting off each punctuation symbol as its own token.  This allows us to use the same embedding for a word regardless of whether it has adjacent punctuation.  For fairness in comparison, we ensure that our tokenizer preserves all information from the original character sequence (i.e., it is reversible).  The exact procedure---which is simple and language-agnostic---is described in \cref{sec:tokenizer}, with accompanying code.

The results for the MWC are shown in \cref{tab:results-mwc} in the form of bits per character (bpc).

\subsection{Comparison to baseline models}\label{sec:baselines}

The first baseline model is a purely character-level RNN
language model (\textsc{\bfseries pure-char}).  It is naturally open-vocabulary
(with respect to words; like all models we evaluate, it does assume  a
closed character set).
This baseline reaches by far the worst bpc rate on the held-out sets, perhaps because it works at too short a time scale to capture long-range dependencies. 

A much stronger baseline---as it turns out---is a subword-level RNN language model (\textsc{\bfseries pure-bpe}).  It models a sequence of \emph{subword units}, where each token in the corpus is split into one or more subword units by \emph{Byte Pair Encoding} (BPE), an old compression technique first used for neural machine translation by \citet{SenHadBir16Neural}.  This gives a kind of interpolation between a word-level model and a character-level model.  The set of subword units is finite and determined from training data, but includes all characters in $\Sigma$, making it posssible to explain any novel word in held-out data.  The size of this set can be tuned by specifiying the number of BPE ``merges.''%
\footnote{\label{fn:doubleseg}%
A segmented example sentence from WikiText-2 is ``The\ \ $|$ex$|$os$|$kel$|$eton\ \ $|$is\ \ $|$gener$|$ally\ \ $|$blue''.
Technically we do not model the string, but the specific segmented string chosen by BPE.  Modeling the string would require marginalizing over all possible segmentations (which is intractable to do exactly with a neural language model); more discussion on that in \cref{sec:latent-segmentation}.
}
To our surprise, it is the strongest competitor to our proposed model, even outperforming it on the MWC.
One has to wonder why BPE has not (to our knowledge) been tried previously as an open-vocabulary language model, given its ease of use and general applicability.

Notice, however, that even when \textsc{pure-bpe} performs well as a language model, it does not provide \emph{word embeddings} to use in other tasks like machine translation, parsing, or entailment.  We cannot extract the usual static type embeddings from it, nor is it obvious how to create dynamic per-token embeddings like the \emph{contextualized embeddings} of \citet{PetNeuIyy18Deep}.  Our model allows for both, namely $\ofw{e}$ and $\vec{h}_i$.

Finally, we also compare against the character-aware model of \citet{KawDyeBlu17Learning}, both without (\textbf{HCLM}) and with their additional cache (\textbf{HCLMcache}).  To our knowledge, that model has the best previously known performance on the \emph{raw} (i.e., open-vocab) version of the WikiText-2 dataset, but we see in both \cref{tab:results} and \cref{tab:results-mwc} that our model and the \textsc{pure-bpe} baseline beat it.

\begin{table}
    \centering
    \begin{tabular}{l||ccc|c||c}
        \textit{\ \ WikiText-2} & \multicolumn{4}{c||}{\textit{dev}} & \textit{test} \\
        \footnotesize types w/ count & \footnotesize 0 & \footnotesize \!\![1, 100)\!\! & \footnotesize \!\![100; $\infty$)\!\! & \footnotesize all & \footnotesize all \\[-.2em]
        \scriptsize \ \ \# of such tokens &\scriptsize 7116 &\scriptsize 47437 &\scriptsize 163077 &\scriptsize $\sum$ & \scriptsize $\sum$ \\\hline\hline
        \textsc{pure-char}       & \textbf{3.89} & 2.08          & 1.38          & 1.741         & 1.775         \\
        \textsc{pure-bpe}        & 4.01          & 1.70          & \textbf{1.08} & 1.430         & 1.468         \\\hline
        \textsc{only-reg} & 4.37 & 1.68 & 1.10 & 1.452 & 1.494 \\
        \textsc{sep-reg}  & 4.17 & 1.65 & 1.10 & 1.428 & 1.469 \\
        \textsc{no-reg}   & 4.14 & 1.65 & 1.10 & 1.426 & 1.462 \\\hline
        \textsc{1gram}    & 5.09 & 1.73 & 1.10 & 1.503 & 1.548 \\
        \textsc{uncond}   & 4.13 & 1.65 & 1.10 & 1.429 & 1.468 \\
        \textsc{full}     & 4.00 & \textbf{1.64} & 1.10 & \textbf{1.416} & \textbf{1.455} \\\hline
        \hline
        HCLM                 & -- & -- & -- & 1.625 & 1.670 \\
        HCLMcache            & -- & -- & -- & 1.480 & 1.500 \\
    \end{tabular}
    \caption{Bits per character (lower is better) on the dev and test set of \textbf{WikiText-2} for our model and baselines, where \textsc{full} refers to our main proposed model
    and HCLM and HCLMcache refer to \citet{KawDyeBlu17Learning}'s proposed models.
    All our hybrid models use a vocabulary size of 50000, \textsc{pure-bpe} uses 40000 merges (both tuned from \cref{fig:vocab-dependence}).
    All pairwise differences except for those between \textsc{pure-bpe}, \textsc{uncond}, and \textsc{sep-reg} are statistically significant (paired permutation test over all 64 articles in the corpus, $p < 0.011$).
    }
    \label{tab:results}
\end{table}

\subsection{Analysis of our model on WikiText-2}\label{sec:analysis}

\subsubsection{Ablating the training objective}

How important are the various influences on $p_\mathrm{spell}$?  Recall that $p_\mathrm{spell}$ is used to relate embeddings of in-vocabulary types to their spellings at training time.  We can omit this \emph{regularization} of in-vocabulary embeddings by dropping the second factor of the training objective, \cref{eqn:ovloss}, which
gives the \textsc{\bfseries no-reg} ablation.  $p_\mathrm{spell}$ is also trained explicitly to spell out \textsc{unk} tokens, which is how it will be used at test time.  Omitting this part of the training by dropping the fourth factor gives the \textsc{\bfseries only-reg} ablation.

We can see in \cref{tab:results} that neither \textsc{no-reg} nor \textsc{only-reg} performs too well (no matter the vocabulary size, as we will see in \Cref{fig:vocab-dependence}).  That is, the spelling model benefits from being trained on both in-vocabulary types and \textsc{unk} tokens.

To tease apart the effect of the two terms, we evaluate what happens if we use two separate spelling models for the second and fourth factors of \cref{eqn:ovloss}, giving us the \textsc{\bfseries sep-reg} ablation.
Now the in-vocabulary words are spelled from a different model and do not influence the spelling of \textsc{unk}.%
\footnote{Though this prevents sharing statistical strength, it might actually be a wise design if \textsc{unk}s are in fact spelled differently (e.g., they tend to be long, morphologically complex, or borrowed).}

Interestingly, \textsc{sep-reg} does not perform better than \textsc{no-reg} (in \cref{fig:vocab-dependence} we see no big difference), suggesting that it is not the ``smoothing'' of embeddings using a speller model that is responsible for the improvement of \textsc{full} over \textsc{no-reg}, but the benefit of training the speller on more data.\footnote{All this, of course, is only evaluated with the hyperparameters chosen for \textsc{full}. Retuning hyperparameters for every condition might change these results, but is infeasible.}

\subsubsection{Speller architecture power}

We also compare our full model (\textsc{full}) against two ablated versions that simplify the spelling model:
\textsc{\bfseries 1gram}, where $p(\ofw{\sigma}) \propto \prod_{i=1}^{|\ofw{\sigma}|} q(\ofw{\sigma}_i)$ (a learned unigram distribution $q$ over characters instead of an RNN) and
\textsc{\bfseries uncond}, where $p(\ofw{\sigma}) \propto p_{\text{spell}}(\ofw{\sigma} \mid \vec{0})$, (the RNN character language model, but without conditioning on a word embedding).

In \cref{tab:results}, we clearly see that as we go from \textsc{1gram} to \textsc{uncond} to \textsc{full}, the speller's added expressiveness improves the model.

\subsubsection{Rare versus frequent words}

It is interesting to look at bpc broken down by word frequency,%
\footnote{We obtain the number for each frequency bin by summing the contextual log-probabilities of the tokens whose types belong in that bin, and dividing by the number of characters of all these tokens.  (For the \textsc{pure-char} and \textsc{pure-bpe} models, the log-probability of a token is a sum over its characters or subword units.)}$^,$\footnote{Low bpc means that the model can predict the tokens in this bin from their \emph{left} contexts.  It does not also assess whether the model makes good use of these tokens to help predict their right contexts.}
shown in \cref{tab:results}.  The first bin contains (held-out tokens of) words that were never seen during training, the second contains words that were only rarely seen (about half of them in $\mathcal{V}$), and the third contains frequent words.
Unsurprisingly, rarer words generally incur the highest loss in bpc, although of course their lower frequency does limit the effect on the overall bpc.

On the frequent words, there is hardly any difference among the several models---they can all memorize frequent words---except that the \textsc{pure-char} baseline performs particularly badly.  Recall that \textsc{pure-char} has to re-predict the spelling of these often irregular types each time they occur.  Fixing this was the original motivation for our model.

On the infrequent words, \textsc{pure-char} continues to perform the worst.  Some differences now emerge among the other models, with our \textsc{full} model winning.  Even the ablated versions of \textsc{full} do well, with 5 out of our 6 beating both baselines.  The advantage of our systems is that they create lexical entries that memorize the spellings of all in-vocabulary training words, even infrequent ones that are rather neglected by the baselines.

On the novel words, our 6 systems have the same relative ordering as they do on the infrequent words.  The surprise in this bin is that the baseline systems do extremely well, with \textsc{pure-bpe} nearly matching \textsc{full} and \textsc{pure-char} beating it, even though we had expected the baseline models to be too biased toward predicting the spelling of frequent words.  Note, however, that $p_{\text{spell}}$ uses a weaker LSTM than $p_{\text{LM}}$ (fewer nodes and different regularization), which may explain the difference.

\subsubsection{Vocabulary size as a hyperparameter}\label{sec:vocab-size}

\begin{figure}
    \hspace*{-.5em}
    \begin{adjustbox}{width=\columnwidth}
        \definecolor{brewer1}{cmyk}{0,.32,.73,.05}
        \definecolor{brewer2}{cmyk}{.22,.16,0,.24}
        \begin{tikzpicture}
            \begin{axis}[
                    axis x line*=bottom, axis y line*=left,
                    xlabel={vocabulary size},
                    ylabel={bpc on the WikiText-2 dev set},
                    scaled ticks={false},
                    tick label style={/pgf/number format/fixed},
                    xmajorgrids,
                    ymajorgrids,
                    xminorgrids,
                    yminorgrids,
                    minor tick num=4,
                    xtick={0,50000,100000},
                    xmin=0, xmax=75000,
                    ymin=0.0, ymax=1.75,
                    width=12em, height=18em,
                    legend style={at={(0.5,1.18)},anchor=north,draw=none,fill=none,column sep=1em,name=legend},
                    legend columns=1,
                    legend cell align=left
                    ]
                \addplot[draw=brewer1,very thick] coordinates {
                    (5000,1.606)
                    (10000,1.517)
                    (20000,1.446)
                    (30000,1.425)
                    (40000,1.417)
                    (45000,1.418)
                    (50000,1.416)
                    (55000,1.426)
                    (60000,1.424)
                    (65000,1.431)
                    (70000,1.442)
                    (75000,1.496)
                    };
                \addplot[draw=brewer1,densely dashdotted, thick] coordinates {
                    (5000,0.947128205128205)
                    (10000,1.03967992766727)
                    (20000,1.10914772727273)
                    (30000,1.15592485549133)
                    (40000,1.18403094777563)
                    (45000,1.19583752417795)
                    (50000,1.20195348837209)
                    (55000,1.22032692307692)
                    (60000,1.22645086705202)
                    (65000,1.23910344827586)
                    (70000,1.25797333333333)
                    (75000,1.29324542124542)
                    };
                \node[color=brewer1,xshift=4em,yshift=8.7em,rotate=10]{\scriptsize $p_{\mathrm{LM}}(\cdot)$ only};
            \end{axis}
            \begin{axis}[
                    xshift=10em,
                    axis x line*=bottom, axis y line*=right,
                    xlabel={vocabulary size},
                    scaled ticks={false},
                    tick label style={/pgf/number format/fixed},
                    xmajorgrids,
                    ymajorgrids,
                    yminorgrids,
                    minor tick num=3,
                    xmin=5000, xmax=75000,
                    ymin=1.4, ymax=1.5,
                    width=17em, height=18em,
                    legend style={at={(0.12, 1.24)},anchor=north,draw=none,fill=none,column sep=1em,name=legend},
                    legend columns=3,
                    legend cell align=left
                    ]
                \addplot[draw=brewer1,ultra thick] coordinates { 
                    (5000,1.606)
                    (10000,1.517)
                    (20000,1.446)
                    (30000,1.425)
                    (40000,1.417)
                    (45000,1.418)
                    (50000,1.416)
                    (55000,1.426)
                    (60000,1.424)
                    (65000,1.431)
                    (70000,1.442)
                    (75000,1.496)
                    };
                \addplot[draw=brewer2] coordinates { 
                    (6849,1.498)
                    (12824,1.471)
                    (24434,1.436)
                    (35670,1.433)
                    (46723,1.43)
                    (57643,1.433)
                    (68593,1.443)
                    (76996,1.451)
                    };
                \addplot[draw=gray,densely dashed, thick] coordinates { 
                    (1000,1.813)
                    (20000,1.451)
                    (30000,1.429)
                    (40000,1.425)
                    (50000,1.426)
                    (60000,1.439)
                    (70000,1.469)
                    (76132,1.713)
                    };
                \addplot[draw=gray,densely dotted, thick] coordinates { 
                    (20000,1.540)
                    (40000,1.460)
                    (50000,1.452)
                    (60000,1.451)
                    (70000,1.457)
                    (76132,1.537)
                    };
                \addplot[draw=gray,densely dashdotted, thick] coordinates { 
                    (10000,1.522)
                    (20000,1.450)
                    (30000,1.424)
                    (40000,1.423)
                    (50000,1.428)
                    (60000,1.439)
                    (70000,1.466)
                    (76132,1.712)
                    };
                \legend{\textsc{full},\textsc{pure-bpe},\textsc{no-reg},\textsc{only-reg},\textsc{sep-reg}}
            \end{axis}
            \draw[color=black!70] (0.0, 4.07) -- (2.85, 4.07) -- (3.25, 4.75) -- (7.915, 4.75); 
            \draw[color=black!70] (0.0, 3.80) -- (2.85, 3.80) -- (3.25, 0.00) -- (7.915, 0.00); 
            \draw[color=black!70] (3.55, 4.75) -- (3.52, 0.0);
        \end{tikzpicture}
    \end{adjustbox}
    \caption{Bits-per-character (lower is better) on WikiText-2 dev data as a function of vocabulary size. Left: The total cross-entropy is dominated by the third factor of \cref{eqn:ovloss}, $p_{\mathrm{LM}}$, the rest being its fourth factor. Right (zoomed in): baselines.}
    \label{fig:vocab-dependence}
\end{figure}
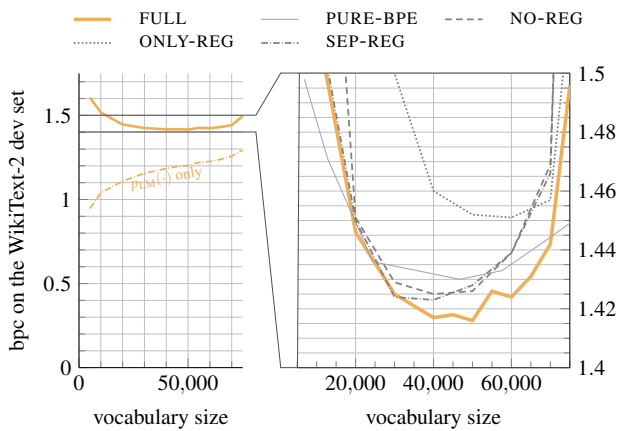

In \cref{fig:vocab-dependence} we see that the size of the vocabulary---a hyperparameter of both the \textsc{pure-bpe} model (indirectly by the number of merges used%
) and our \textsc{full} model and its ablations---does influence results noticeably.  There seems to be a fairly safe plateau when selecting the 50000 most frequent words (from the raw  WikiText-2 vocabulary of about 76000 unique types), which is what we did for \cref{tab:results}.  Note that at \emph{any} vocabulary size, both models perform far better than \textsc{pure-char}, whose bpc of 1.775 is far above the top of the graph.

\Cref{fig:vocab-dependence} also shows that as expected, the loss of the \textsc{full} model (reported as bpc on the entire dev set) is made up mostly of the cross-entropy of $p_\mathrm{LM}$.  This is especially so for larger vocabularies, where very few \textsc{unk}s occur that would have to be spelled out using $p_\mathrm{spell}$.

\subsection{Results on the multilingual corpus}

\newcommand{\hidenumber}[1]{}
\newcommand{\justsayk}[1]{k}
\begin{table*}
    \centering
    \begin{tabular}{cr||cc|cc|cc|cc|cc|cc|cc}
        \multirow{2}{*}{\textit{MWC\hspace*{-2em}}} && \multicolumn{2}{c|}{\textit{en}} & \multicolumn{2}{c|}{\textit{fr}} & \multicolumn{2}{c|}{\textit{de}} & \multicolumn{2}{c|}{\textit{es}} & \multicolumn{2}{c|}{\textit{cs}} & \multicolumn{2}{c|}{\textit{fi}} & \multicolumn{2}{c}{\textit{ru}} \\
        && \footnotesize dev & \footnotesize test & \footnotesize dev & \footnotesize test & \footnotesize dev & \footnotesize test & \footnotesize dev & \footnotesize test & \footnotesize dev & \footnotesize test & \footnotesize dev & \footnotesize test & \footnotesize dev & \footnotesize test \\\hline\hline
        \multirow{5}{*}{\rotatebox{90}{\footnotesize space-split}}
        & \footnotesize\em\!\!\!\rule{0pt}{1em}\#{}types$\to{}^{\text{merges}}{\mskip -5mu/\mskip -3mu}_{\text{vocab}}$\!\! & \multicolumn{2}{c|}{\footnotesize\em 195\justsayk{572} $\to$ 60k} & \multicolumn{2}{c|}{\footnotesize\em 166\justsayk{865} $\to$ 60k} & \multicolumn{2}{c|}{\footnotesize\em 242\justsayk{481} $\to$ 60k} & \multicolumn{2}{c|}{\footnotesize\em 162\justsayk{402} $\to$ 60k} & \multicolumn{2}{c|}{\footnotesize\em 174\justsayk{638} $\to$ 60k} & \multicolumn{2}{c|}{\footnotesize\em 191\justsayk{599} $\to$ 60k} & \multicolumn{2}{c}{\footnotesize\em 244\justsayk{185} $\to$ 60k} \\
        \cline{2-16}
        & \textsc{pure-bpe}\rule{0pt}{0.95em}\!\! & \!\textcolor{gray}{1.50\hidenumber{6}}\! & \!1.439\! & \!\textcolor{gray}{1.40\hidenumber{7}}\! & \!1.365\! & \!\textcolor{gray}{1.49\hidenumber{8}}\! & \!1.455\! & \!\textcolor{gray}{1.46\hidenumber{7}}\! & \!1.403\! & \!\textcolor{gray}{1.92\hidenumber{8}}\! & \!1.897\! & \!\textcolor{gray}{1.73\hidenumber{3}}\! & \!1.685\! & \!\textcolor{gray}{1.68\hidenumber{6}}\! & \!1.643\! \\
        & \textsc{full}\!\!     & \!\textcolor{gray}{1.57\hidenumber{5}}\! & \!1.506\! & \!\textcolor{gray}{1.48\hidenumber{2}}\! & \!1.434\! & \!\textcolor{gray}{1.66\hidenumber{2}}\! & \!1.618\! & \!\textcolor{gray}{1.53\hidenumber{7}}\! & \!1.469\! & \!\textcolor{gray}{2.27\hidenumber{9}}\! & 2.240 & \!\textcolor{gray}{1.93\hidenumber{9}}\! & \!1.896\! & \!\textcolor{gray}{2.00\hidenumber{3}}\! & \!1.969\! \\
        & HCLM\!\!              & \!\textcolor{gray}{1.68\hidenumber{3}}\! & \!1.622\! & \!\textcolor{gray}{1.55\hidenumber{3}}\! & \!1.508\! & \!\textcolor{gray}{1.66\hidenumber{6}}\! & \!1.641\! & \!\textcolor{gray}{1.61\hidenumber{7}}\! & \!1.555\! & \!\textcolor{gray}{2.07\hidenumber{0}}\! & 2.035 & \!\textcolor{gray}{1.83\hidenumber{2}}\! & \!1.796\! & \!\textcolor{gray}{1.83\hidenumber{2}}\! & \!1.810\! \\
        & HCLMcache\!\!        & \!\textcolor{gray}{1.59\hidenumber{1}}\! & \!1.538\! & \!\textcolor{gray}{1.49\hidenumber{9}}\! & \!1.467\! & \!\textcolor{gray}{1.60\hidenumber{5}}\! & \!1.588\! & \!\textcolor{gray}{1.54\hidenumber{8}}\! & \!1.498\! & \!\textcolor{gray}{2.01\hidenumber{0}}\! & \!1.984\! & \!\textcolor{gray}{1.75\hidenumber{4}}\! & \!1.711\! & \!\textcolor{gray}{1.77\hidenumber{7}}\! & \!1.761\! \\\hline\hline
        \multirow{3}{*}{\rotatebox{90}{\!\!\footnotesize tokenize}}
        & \footnotesize\em\!\!\!\rule{0pt}{1em}\#{}types$\to{}^{\text{merges}}{\mskip -5mu/\mskip -3mu}_{\text{vocab}}$\!\! & \multicolumn{2}{c|}{\footnotesize\em 94\justsayk{726} $\to$ 60k} & \multicolumn{2}{c|}{\footnotesize\em 88\justsayk{234} $\to$ 60k} & \multicolumn{2}{c|}{\footnotesize\em 157\justsayk{331} $\to$ 60k} & \multicolumn{2}{c|}{\footnotesize\em 93\justsayk{318} $\to$ 60k} & \multicolumn{2}{c|}{\footnotesize\em 126\justsayk{272} $\to$ 60k} & \multicolumn{2}{c|}{\footnotesize\em 147\justsayk{139} $\to$ 60k} & \multicolumn{2}{c}{\footnotesize\em 166\justsayk{929} $\to$ 60k} \\
        \cline{2-16}
        & \textsc{pure-bpe}\rule{0pt}{0.95em}\!\! & \textbf{\!\textcolor{gray}{1.45\hidenumber{3}}\!} & \textbf{\!1.386\!} & \textbf{\!\textcolor{gray}{1.36\hidenumber{6}}\!} & \textbf{\!1.317\!} & \textbf{\!\textcolor{gray}{1.45\hidenumber{6}}\!} & \textbf{\!1.414\!} & \textbf{\!\textcolor{gray}{1.42\hidenumber{7}}\!} & \textbf{\!1.362\!} & \textbf{\!\textcolor{gray}{1.88\hidenumber{7}}\!} & \textbf{\!1.856\!} & \textbf{\!\textcolor{gray}{1.70\hidenumber{6}}\!} & \textbf{\!1.652\!} & \textbf{\!\textcolor{gray}{1.63\hidenumber{7}}\!} & \textbf{\!1.598\!} \\
        & \textsc{full}\!\! & \textbf{\!\textcolor{gray}{1.45\hidenumber{2}}\!} & \textbf{\!1.387\!} & \textbf{\!\textcolor{gray}{1.36\hidenumber{9}}\!} & \textbf{\!1.319\!} & \!\textcolor{gray}{1.51\hidenumber{0}}\! & \!1.465\! & \textbf{\!\textcolor{gray}{1.42\hidenumber{8}}\!} & \textbf{\!1.363\!} & \!\textcolor{gray}{1.95\hidenumber{6}}\! & \!1.928\! & \!\textcolor{gray}{1.79\hidenumber{8}}\! & \!1.751\! & \!\textcolor{gray}{1.74\hidenumber{5}}\! & \!1.709\! \\
    \end{tabular}
    \caption{Bits per character (lower is better) on the dev and test sets of the \textbf{MWC} for our model (\textsc{full}) and \citet{KawDyeBlu17Learning}'s HCLM and HCLMcache, both on the space-split version used by \citet{KawDyeBlu17Learning} and the more sensibly tokenized version. Values across all rows are comparable, since the tokenization is reversible and bpc is still calculated w.r.t. the number of characters in the original version.
    All our models did not tune the vocabulary size, but use 60000.
    }
    \label{tab:results-mwc}
\end{table*}

We evaluated on each MWC language using the system and hyperparameters
that we had tuned on WikiText-2 development data.  Even the vocabulary size stayed fixed at 60000.\footnote{Bigger vocabularies require smaller batches to fit our GPUs, so changing the vocabulary size would have complicated fair comparisons across methods and languages, as the batch size has a large influence on results.  However, the optimal vocabulary size is presumably language- and dataset-dependent.}

Frustratingly, lacking tuning to MWC, we do not outperform our own (novel) BPE baseline on MWC.  We perform at most equally well, even when leveling the playing field through proper tokenization (\cref{sec:mwc}).
Nevertheless we outperform the best model of \citet{KawDyeBlu17Learning} on most datasets, even when using the space-split version of the data (which, as explained in \cref{sec:mwc}, hurts our models).

Interestingly, the datasets on which we lose to \textsc{pure-bpe} are Czech, Finnish, and Russian---languages known for their morphological complexity.
Note that \textsc{pure-bpe} greatly benefits from the fact that these languages have a concatenative morphological system unlike Hebrew or Arabic.
Explicitly incorporating morpheme-level information into our \textsc{full} model might be useful (cf. \citet{MatNeuDye18Using}).
Our present model or its current hyperparameter settings (especially the vocabulary size) might not be as language-agnostic as we would like.

\subsection{What does the speller learn?}\label{sec:qualitative-eval}

Finally, \cref{tab:pspellsamples} presents non-cherry-picked samples from $p_\mathrm{spell}$, after training our \textsc{full} model on WikiText-2.
$p_\mathrm{spell}$ seems to know how to generate appropriate random forms that appear to have the correct part-of-speech, inflectional ending, capitalization, and even length.

\begin{table}
    \centering
    \begin{tabular}{ll}
        $\sigma(w)$ & $\spelling \sim p_\mathrm{spell}(\cdot \mid e(w))$\\\hline
        \charword{grounded} & \charword{stipped} \\
        \charword{differ} & \charword{coronate} \\
        \charword{Clive} & \charword{Dickey} \\
        \charword{Southport} & \charword{Strigger} \\
        \charword{Carl} & \charword{Wuly} \\
        \charword{Chants} & \charword{Tranquels} \\
        \charword{valuables\ \ \ \ \ \ \ } & \charword{migrations} \\
    \end{tabular}
    \caption{Take an in-vocabulary word $w$ (non-cherry-picked), and compare $\sigma(w)$ to a random spelling $\spelling \sim p_\mathrm{spell}(\cdot \mid e(w))$.}
    \label{tab:pspellsamples}
\end{table}

We can also see how the speller chooses to create forms \emph{in context}, when trying to spell out \textsc{unk} given the hidden state of the lexeme-level RNN.
The model knows \emph{when} and \emph{how} to generate sensible years, abbreviations, and proper names, as seen in the example in the introduction (\cref{sec:introduction}).\footnote{Generated at temperature $T=0.75$ from a \textsc{full} model with $|\mathcal{V}| = 20000$.}
Longer, non-cherry-picked samples for several of our models can be found in \cref{sec:textsamples}.

\section{Related work}\label{sec:previous-work}

\begin{table}[t]
    \centering
    \begin{tabular}{r||c|c}
        & closed-vocab & open-vocab \\\hline\hline
        \begin{minipage}{2.5em}\raggedleft (pure) words\end{minipage} &
            \begin{minipage}{6.5em}\vspace*{.5em}\centering\citet{MikKarBur10Recurrent}, \citet{SunSchNey12LSTM}\vspace*{.5em}\end{minipage} &
            \emph{\color{gray} -impossible-}\\\hline
        \begin{minipage}{2.5em}\raggedleft words + chars\end{minipage} &
            \begin{minipage}{6.5em}\centering\citet{KimJerSon16Character}, \citet{LinDyeBla15Finding}\end{minipage} &
            \begin{minipage}{10em}\vspace*{.5em}\centering\citet{KawDyeBlu17Learning}, \citet{HwaSun17Character}, $\bigstar$\\\vspace*{.5em}\end{minipage}\\\hline
        \begin{minipage}{2.5em}\raggedleft (pure) chars\end{minipage} &
            \emph{\color{gray} -impossible-} &
            \begin{minipage}{10em}\vspace*{.5em}\centering\citet{SutMarHin11Generating}\vspace*{.25em}\end{minipage}\\
    \end{tabular}
    \caption{Contextualizing this work ($\bigstar$) on two axes}
    \label{tab:contextualizing}
\end{table}

Unlike most previous work, we try to \emph{combine} information about words and characters to achieve open-vocabulary modeling.
The extent to which previous work achieves this is as shown in \cref{tab:contextualizing} and explained in this section.

\citet{MikKarBur10Recurrent} first introduced a purely word-level (closed-vocab) RNN language model (later adapted to LSTMs by \citet{SunSchNey12LSTM}).
\citet{SutMarHin11Generating} use an RNN to generate pure character-level sequences, yielding an open-vocabulary language model, but one that does not make use of the existing word structure.

\citet{KimJerSon16Character} and \citet{LinDyeBla15Finding} first combined the two layers by deterministically constructing word embeddings from characters (training the embedding function on tokens, not types, to ``get frequent words right''---ignoring the issues discussed in \cref{sec:words-characters-types-tokens}). Both only perform language modeling with a closed vocabulary and thus use the subword information only to improve the estimation of the word embeddings (as has been done before by \citet{SanZad14Learning}).

Another line of work
instead augments a character-level RNN with word-level ``impulses.''
Especially noteworthy is the work of \citet{HwaSun17Character}, who describe an architecture in which character-level and word-level models run in parallel from left to right and send vector-valued messages to each other.  The word model sends its hidden state to the character model, which generates the next word, one character at a time, and then sends its hidden state back to update the state of the word model.
However, as this is another example of \emph{constructing} word embeddings from characters, it again overemphasizes learning frequent spellings (\cref{sec:typetoken}).

Finally, the most relevant previous work is the (independently developed) model of \citet{KawDyeBlu17Learning}, where each word has to be ``spelled out'' using a character-level RNN if it cannot be directly copied from the recent past.
As in \citet{HwaSun17Character}, there is no fixed vocabulary, so words that have fallen out of the cache have to be re-spelled.
Our hierarchical generative story---specifically, the process that generates the lexicon---handles the re-use of words more gracefully.  Our speller can then focus on representative phonotactics and morphology of the language instead of generating frequent function words like \charword{the} over and over again.
Note that the use case that Kawakami et al.\@ originally intended for their cache, the copying of highly infrequent words like \charword{Noriega} that repeat on a very local scale \citep{Chu00Empirical}, is not addressed in our model, so adding their cache module to our model might still be beneficial.

Less directly related to our approach of improving language models is the work of \citet{BhaGutEis16Morphological}, who similarly realize that placing priors on word embeddings is better than compositional construction, and \citet{PinGutEis17Mimicking}, who prove that the spelling of a word shares information with its embedding.

Finally, in the highly related field of machine translation, \citet{LuoMan16Achieving} before the re-discovery of BPE proposed an open-vocabulary neural machine translation model in which the prediction of an \textsc{unk} triggers a character-level model as a kind of ``backoff.'' We provide a proper Bayesian explanation for this trick and carefully ablate it (calling it \textsc{no-reg}), finding that it is insufficient, and that training on types (as suggested by far older research) is more effective for the task of language modeling.

\section{Conclusion}\label{sec:conclusion-future}

We have presented a generative two-level open-vocabulary language model that can memorize spellings and embeddings of common words, but can also generate new word types in context, following the spelling style of in- and out-of-vocabulary words.  This architecture is motivated by linguists' ``duality of patterning.'' It resembles prior Bayesian treatments of type reuse, but with richer (LSTM) sequence models.

We introduced a novel, surprisingly strong baseline, beat it by tuning our model, and carefully analyzed the performance of our model, baselines, and a variety of ablations on multiple datasets.
The conclusion is simple: pure character-level modeling is not appropriate for language, nor required for an open vocabulary.
Our ablations show that the generative story our model is based on is superior to distorted or simplified models resembling previous ad-hoc approaches.

In future work, our approach could be used in other generative NLP models that use word embeddings.  Our spelling model relates these embeddings to their spellings, which could be used to regularize embeddings of rare words (using the speller loss as another term in the generation process), or to infer embeddings for unknown words to help make closed-vocabulary models open-vocabulary.
Both are likely to be extremely helpful in tasks like text classification (e.g., sentiment), especially in low-resource languages and domains.

\section*{Acknowledgments}
This material is based upon work supported by the National Science Foundation under Grant No.\@ 1718846.
Part of the research made use of computational resources at the Maryland Advanced Research Computing Center (MARCC).
The authors would like to thank Benjamin Van Durme and Yonatan Belinkov for helpful suggestions on experiments, and Annabelle Carrell as well as the anonymous reviewers for their suggestions on improving the exposition.

{
\fontsize{9.25pt}{10.0pt}\selectfont
\bibliography{inlm}
\bibliographystyle{aaai}
}

\appendix%
\clearpage%
\appendix%
\appendixpage%
\fontsize{9.25pt}{10.0pt}\selectfont

\section{Training the joint model}\label{sec:joint-model}

\subsection{Adding all losses}

We optimize the parameters $\theta$ of $p_\mathrm{LM}$ and $p_\mathrm{spell}$ by maximizing the joint probability of the parameters and our training data, as given in \cref{eqn:ovloss} above.
We regard the training data as one long tokenized sequence, with each line break having been replaced by an end-of-sentence \textsc{eos} token (thus, $\textsc{eos} \in \mathcal{V}$).
This is standard practice in the language modeling community at least since \citet{MikKarBur10Recurrent},
as it allows the model to learn long-range content-sensitive dependencies.

Computing \cref{eqn:ovloss} is easy enough.  First, we run the spelling model on the in-vocabulary embedding-spelling pairs $\{\ofw{e}, \ofw{\sigma}\}_{w \in \mathcal{V} \setminus \{\textsc{unk}, \textsc{eos}\}}$,\footnote{Spellings exceeding 20 characters are omitted to speed up the training process.} to compute the second factor of \cref{eqn:ovloss}.
To get the third factor, we run the lexeme-level RNN over the sequence $w_1 \cdots w_n$, as it is defined in \cref{sec:spelling-unks}, recording the hidden state $\vec{h}_i$ for each time step $i$ where $w_i = \textsc{unk}$.  Finally, we can use the actual spelling and recorded hidden states at these timesteps $\spelling_i$ (along with the in-vocabulary lexemes for which we have embeddings) to compute the fourth factor of \cref{eqn:ovloss}.

We maximize \cref{eqn:ovloss} by stochastic gradient descent on its negative logarithm \citep{RobMon51Stochastic}, where we sample the stochastic gradient as follows.

\subsection{Batching}\label{sec:batching}

As it is impractical to compute the gradient of the language model likelihood (the log of the third and fourth factors) on a single string of millions of tokens, the standard practice in language modeling\footnote{As noted by \citet{MikKarBur10Recurrent} and reflected even in reference implementations of frameworks like PyTorch: {\scriptsize \url{https://github.com/pytorch/examples/tree/0.3/word_language_model}}} is to obtain an approximate stochastic gradient via stochastically truncated backpropagation-through-time, which reduces memory requirements and allows more frequent updates.  We also break the training string into 40 strings, so that each stochastic gradient sample is the ``minibatch'' sum of 40 gradients that can be computed in parallel on a GPU.  The resulting gradient is the gradient of the log-probability of the tokens in the minibatch; to obtain a stochastic gradient for the entire language model likelihood, we divide by the the number of tokens in the minibatch and multiply by the number of tokens in the corpus.

On every \nth{100} step, we augment the above stochastic gradient by adding a sample of the stochastic gradient of the log of the second factor, which is obtained by summing over a minibatch of 2000 lexemes from $\mathcal{V}$, and multiplying by $|\mathcal{V}|/2000$.  This sample is further upweighted by a factor of 100 to compensate for its infrequency.  Because these steps are so infrequent, our model's training time is only negligibly worse than that of the closed-vocab language model of \citet{MerKesSoc17Regularizing} (even though we still have to account for all \textsc{unk} spellings (the fourth factor)).

Each gradient step also includes a ``weight decay'' operation that shrinks all parameters of the model toward 0, which corresponds to augmenting the stochastic gradient by adding the gradient of the log of the first factor, $\nabla \log p(\theta) \propto \theta$.

\section{Hyperparameters}\label{sec:hyperparameters}

\subsection{The lexeme-level RNN}

The lexeme-level RNN is a three-layer Averaged Stochastic Gradient Descent with Weight Dropped LSTM \citep{MerKesSoc17Regularizing}, using 400-dimensional word embeddings and 1150-dimensional hidden states.
The size of the vocabulary is set to 60000.
As described in \cref{sec:batching}, we use batching for the lexeme-level RNN.
However, while we generally copy the hyperparameters that \citet{MerKesSoc17Regularizing} report as working best for the WikiText-2 dataset (see \cref{sec:wikitext2}), we have to change the batch size from 80 down to 40 and limit the sequence length (which is sampled from the normal distribution $\mathcal{N}(70, 5)$ with probability $0.95$ and $\mathcal{N}(35, 5)$ with probability $0.05$) to a maximum of 80 (as \citet{MerKesSoc17Regularizing} advise for training on K80 GPUs) --- we find neither to have meaningful impact on the scores reported for the tasks in \citet{MerKesSoc17Regularizing}.

\subsection{The speller RNN}\label{sec:speller-hyperparams}

For the speller RNN, we implement a three-layer vanilla LSTM model following the definition in \cref{sec:speller-model} with 100 hidden units and 5-dimensional embeddings for each character, dropout of $0.2$ for the network and $0.5$ for the word embeddings, a factor of $1$ for the nuclear norm loss, and weight decay $1.2e\text{-}6$.
We note that the a smaller network (with either less layers or fewer hidden states) noticeably underperforms; the other parameters seem less important.
As mentioned in \cref{sec:batching}, the batches of in-vocabulary lexemes that we predict contain 1500 lexemes and are sampled from the set of word types on every \nth{50} batch of the lexeme-level RNN for WikiText-2 and every \nth{100} batch for the MWC.

\subsection{Joint Optimization}

The lexeme-level RNN and the embeddings {\large $\ofw{e}$} are first trained using SGD, then using Averaged SGD (as described in \citet{MerKesSoc17Regularizing}); the speller RNN is trained using SGD. Both use a constant learning rate of 30.
All gradients are clipped to 0.25.

\subsection{The \textsc{pure-char} RNN baseline}

For the \textsc{pure-char} baseline we again use the AWD-LSTM \citep{MerKesSoc17Regularizing}, but we adapt:
\begin{description}
    \item[Batch size] 20
    \item[Mean sequence length] 100
    \item[Dropout (all dropouts)] 0.1
    \item[Token embedding size] 10
    \item[Learning rate] 5
    \item[Epochs] 150 (convergence)
    \item[Vocabulary size] 145 (corresponding to all characters that appear at least 25 times)
\end{description}

All these parameters were tuned on the development data to ensure that the baseline is fair.

\subsection{The \textsc{pure-bpe} RNN baseline}

For our BPE baseline we use the scripts provided by \citet{SenHadBir16Neural} to learn encodings and split words. We perform 50k merges to yield a vocabulary that is comparable to the 60k vocabulary we use for our model on WikiText-2.

Because most units that are produced by BPE splitting are words (or very large subword units), we use the AWD-LSTM-LM with the default parameters of \citet{MerKesSoc17Regularizing}.

\section{The nuclear norm as a regularization tool}\label{sec:nuclearnorm}

\subsection{What is the nuclear norm and why do we want it?}

As explained in \cref{sec:speller-model}, giving the spelling model access to high-dimensional word embeddings risks overfitting.
We hypothesize that the information that the spelling model actually needs to model phonotactics and morphology can be represented in a much lower-dimensional space.
A straightforward idea would be to ``compress'' the word embeddings into such a low-dimensionality subspace using some linear transformation, but this leaves us with yet more parameters to train and one more potentially difficult-to-optimize hyperparameter (the number of dimensions in this lower subspace).
We instead opt for a ``soft rank reduction'' by regularizing the part of the input matrix of the speller RNN that receives the word embeddings towards a low rank, allowing the model to overcome the regularization, if necessary.

The nuclear norm (or \emph{trace norm}) of a matrix $A$ of shape $m \times n$ is defined:
\[
    ||A||_* = \mathrm{trace}(\sqrt{A^* A}) = \sum_{i=1}^{\min\{m,n\}} \!\!\! \sigma_i(A),
\]
where $\sigma_i(A)$ denotes\footnote{The use of $\sigma$ in this paragraph differs from its use in the main paper.} the $i$-th \emph{singular value} of $A$.
The trace norm is a specific version of the \emph{Schatten $p$-norm}:
\[
    ||A||_p = \Big( \sum_{i=1}^{\min\{m,n\}} \!\!\! \sigma_i(A)^p \Big)^\frac{1}{p},
\]
obtained by setting $p=1$, i.e., it is the $\ell_1$-norm of the singular values of a matrix.

How is this related to low-rankness?
If $A$ is of rank $r < \min\{m,n\}$, then $\sigma_i(A) = 0$ for any $i > r$.
Thus we can see that minimizing the trace norm of a matrix not only minimizes the magnitude of individual entries, but also acts as a proxy for rank reduction.%
\footnote{Actual rank reduction would happen if singular values would indeed become 0, but we will be content with getting them close to 0.}

\subsection{How do we calculate it?}

We can obtain Schatten $p$-norms of matrices by computing a \emph{singular value decomposition} (SVD) for them: given a matrix $A$ of shape $m \times n$, we can factorize $A = U \Sigma V^*$, such that $U$ and $V$ are \emph{unitary} (\emph{orthogonal}, if $A \in \mathbb{R}^{m \times n}$) matrices of shape $m \times \min\{m,n\}$ and $\Sigma$\footnote{We apologize for yet another notational clash, but note that this one like the last only happens in this appendix section.} is a \emph{diagonal} matrix of shape $\min\{m,n\} \times \min\{m,n\}$ containing exactly the singular values of $A$ that we need to compute any Schatten $p$-norm and $||A||_* = \mathrm{trace}(\Sigma)$.

As a (sub-)gradient we simply use $U^* V$ (from version 0.4 on PyTorch supports backpropagation through SVD, but as our code uses version 0.3, we can not yet make use of that feature).
Note that this approach does not allow us to get ``true'' 0 singular values, to do that we would have to perform proximal gradient steps, complicating the implementation.
We will make do with getting close to 0 for our model.

\section{Unexpected latent variables}\label{sec:latent-mess}

In both our model and the \textsc{pure-bpe} baseline we end up with a surprising (and not easy to spot) latent variable when modeling a string: our model can generate an in-vocabulary word two ways (as mentioned in \cref{fn:doublegen}) and the \textsc{pure-bpe} baseline can output a string in a number of different segmentations (\cref{fn:doubleseg}).

\subsection{\textsc{unk}-ness of an in-vocabulary word}\label{sec:latent-unkness}

Technically, a spelling $\spelling$ that is actually associated with a lexeme in $\mathcal{V} \setminus \{\textsc{unk}\}$, i.e., for which there is some $\circnum{i} \in \mathcal{V} \setminus \{\textsc{unk}\}$, could now be generated two ways: as $\sigma(\circnum{i})$ and as $p(\spelling \mid \vec{h})$, resulting in a latent-variable model (which would make inference much more complicated). To remedy this, we could explicitly set $p(\spelling \mid \vec{h}) = 0$ for any $\spelling$ for which there is some $\circnum{i} \in \mathcal{V} \setminus \{\textsc{unk}\}$ with $\sigma(\circnum{i}) = \spelling$ and any $\vec{h}$, which would require us to renormalize:
\[p(\spelling \mid \vec{h}) = \frac{p_\mathrm{spell}(\spelling \mid \vec{h})}{1 - \sum_{w \in \mathcal{V}} p_\mathrm{spell}(\ofw{\sigma} \mid \vec{h})))}.\]
In practice, however, the denominator is very close to 1 (because our regularization prevents the speller from overfitting on the training words), so we ignore this issue in our implementation. Since this can only result in an \emph{overestimation} of final bpc values, our evaluation can only make our method look worse than it should.

\subsection{\textsc{pure-bpe} can output a string in any segmentation}\label{sec:latent-segmentation}

Take the example sentence ``The exoskeleton is greater''.
The commonly used reference BPE implementation of \citet{SenHadBir16Neural} is entirely deterministic in the way it segments words into subword units, so let's us assume this ``canonical'' segmentation is ``The\, ex$|$os$|$kel$|$eton\, is\, great$|$er''.

This segmentation is the result of merges between smaller units.
A model over subword units could however have produced the exact same sentence, simply by predicting such smaller units, e.g.:
``T$|$he\, ex$|$os$|$kel$|$eton\, is\, gr$|$eat$|$er'' or even ``T$|$h$|$e\, e$|$x$|$o$|$s$|$k$|$e$|$l$|$e$|$t$|$o$|$n\, i$|$s\, g$|$r$|$e$|$a$|$t$|$e$|$r''.
We can see that what we are actually modeling in our \textsc{pure-bpe} baseline is not the probability of a string, but the probability of a string and its ``best'' segmentation (i.e., the one that is actually predicted by our generative model).
This ambiguity has been used as a regularization before by \citet{Kud18Subword}, but in our application we do need the total probability of the string, so we would have to marginalize over \emph{all possible segmentations}!

This is obviously absolutely intractable and so we are left with no choice but to simply approximate the probability of all segmentations as the probability of the best segmentation.\footnote{Alternatively, we could try restricting the model to only produce ``canonical'' segmentations, but this would require significant engineering logic that would be hard to parallelize and transfer onto GPUs for the caluclation of the vocabulary softmax.}

\section{A simple, reversible, language-agnostic tokenizer}\label{sec:tokenizer}

\newcommand{\mergesymbol}{%
\hspace{.1em}%
\begin{tikzpicture}[y=1.5pt, x=1.5pt, yscale=-1]
    \begin{scope}[fill=black,line join=miter,line cap=butt,line width=0.212pt]
        \path[fill,line width=0.212pt] (-17.1048,125.2070) -- (-14.3339,125.2070) --
            (-14.7993,124.3943) -- (-14.7728,124.3943) -- (-13.9215,125.2383) --
            (-13.9215,124.4546) -- (-13.5308,124.4546) -- (-13.5308,126.3501) --
            (-13.9215,126.3501) -- (-13.9215,125.5663) -- (-14.7728,126.4104) --
            (-14.7993,126.4104) -- (-14.3339,125.5977) -- (-17.1048,125.5977) --
            cycle(-13.5308,123.9096) -- (-16.3017,123.9096) -- (-15.8363,124.7223) --
            (-15.8628,124.7223) -- (-16.7141,123.8782) -- (-16.7141,124.6620) --
            (-17.1048,124.6620) -- (-17.1048,122.7665) -- (-16.7141,122.7665) --
            (-16.7141,123.5503) -- (-15.8628,122.7062) -- (-15.8363,122.7062) --
            (-16.3017,123.5189) -- (-13.5308,123.5189) -- cycle;
    \end{scope}
\end{tikzpicture}%
\hspace{.1em}%
}

\subsection{Universal character categories}

The Unicode standard defines all symbols in use in current computer systems. In it, each symbol is assigned to exactly one ``General category''
, e.g., \textsf{Lu} for ``Letter, Uppercase'', \textsf{Ll} for ``Letter, Lowercase'', \textsf{Sc} for ``Symbol, Currency'', or \textsf{Cc} for ``Other, Control''.

We define the set of ``\emph{weird}'' characters, i.e., characters we want to break the string on as those whose category does not start with \textsf{L} (i.e., letters), with \textsf{M} (i.e., marks like accents), or with \textsf{N} (i.e., numbers), and which are not ``\emph{space}'' either, where ``space'' is defined as a character that Python's \texttt{str.isspace()} method returns true on.\footnote{It would be tempting to use \textsf{Z*}, i.e., the Unicdoe ``Separator'' category, as this third option, but since Python classifies some control characters (i.e., characters in \textsf{Cc}) as spaces, we use this behavior to ensure compatibility with Python whitespace splitting.}

\subsection{Tokenize}

To tokenize a string, we look at each character $c_i$ of the string:
\begin{enumerate}
    \item If it is not \emph{weird}, output it as it is.
    \item If it is weird, we need to split and leave markers for detokenization:
    \begin{enumerate}
        \item If $c_{i-1}$ is not \emph{space} (i.e., we are really introducing a new split before this weird character), output a space and a merge symbol ``\mergesymbol{}''.
        \item Output $c_i$.
        \item If $c_{i+1}$ is not \emph{space} (i.e., we are really introducing a new split after this weird character) \textbf{and} not \emph{weird} (if it is, it will just split itself off from the left context, no need to split now), output a merge symbol ``\mergesymbol{}'' and a space.
    \end{enumerate}
\end{enumerate}

Tokenization thus turns a string like: ``Some of 100,000 households (usually, a minority) ate breakfast.'' into ``Some\;\; of\;\; 100\;\; \mergesymbol{},\mergesymbol{}\;\; 000\;\; households\;\; (\mergesymbol{}\;\; usually\;\; \mergesymbol{},\;\; a\;\; minority\;\; \mergesymbol{})\;\; ate\;\; breakfast\;\; \mergesymbol{}.''.

\subsection{Detokenize}

Again, we look at each character $c_i$ of the string that is to be detokenized:
\begin{enumerate}
    \item If $c_i$ is a space, $c_{i+1}$ is the merge symbol ``\mergesymbol{}'', and $c_{i+2}$ is \emph{weird}, skip ahead to $c_{i+2}$ (i.e., undo a right split).
    \item Otherwise, if $c_i$ is weird, $c_{i+1}$ is the merge symbol ``\mergesymbol{}'', and $c_{i+2}$ is a space, output $c_i$ and move on to $c_{i+3}$ (i.e., undo a left split).
    \item Otherwise, just write out $c_i$ and then continue to $c_{i+1}$.
\end{enumerate}

\subsection{Python implementation}

\begin{figure*}[t]
\begin{lstlisting}[language=python]
  MERGESYMBOL = '§\mergesymbol§'

  
  def is_weird(c):
    return not (unicodedata.category(c)[0] in ['L', 'M', 'N']
                or c.isspace())


  def tokenize(instring):
    for i in range(len(instring)):
      c = instring[i]
      c_p = instring[i-1] if i > 0 else c
      c_n = instring[i+1] if i < len(instring) - 1 else c

      if not is_weird(c):
        stdout.write(c)
      else:
        if not c_p.isspace():
          stdout.write(' ' + MERGESYMBOL)
        stdout.write(c)
        if not c_n.isspace() and not is_weird(c_n):
          stdout.write(MERGESYMBOL + ' ')


  def detokenize(instring):
    i = 0
    while i < len(instring):
      c = instring[i]
      c_p = instring[i-1] if i > 0 else c
      c_n = instring[i+1] if i < len(instring) - 1 else c
      c_pp = instring[i-2] if i > 1 else c
      c_nn = instring[i+2] if i < len(instring) - 2 else c

      if c + c_n == ' ' + MERGESYMBOL and is_weird(c_nn):
        i += 2
      elif is_weird(c) and c_n + c_nn == MERGESYMBOL + ' ':
        stdout.write(c)
        i += 3
      else:
        stdout.write(c)
        i += 1
\end{lstlisting}
\caption{Simplified version of our tokenizer in Python/pseudocode}
\label{fig:tokenize}
\end{figure*}

In summary, the relevant methods are displayed in Python/pseudocode in \cref{fig:tokenize}.

A full and commented implementation can be found at \url{https://sjmielke.com/papers/tokenize}.

\section{Samples from the full model}\label{sec:textsamples}

We show some non-cherrypicked samples\footnote{Truncated to a paragraph to save the trees - peek at the \LaTeX!} from the model for different vocabularies $\mathcal{V}$ and sampling temperatures $T$ (used in the lexeme-level softmax and the speller's character-level softmax both), where in-vocabulary types are printed {\sffamily\color{gray} like this} and newly generated words (i.e., spelled out \textsc{unk}s) are printed \wughighlight{like this}.

Note that the problem explained in \cref{sec:latent-mess} is apparent in these samples: the speller sometimes generates in-vocab words.

\subsection{$|\mathcal{V}| = 60000, T = 0.75$}

{\footnotesize\sffamily\color{gray}
= = = \wughighlight{Comparish} = = =

From late May to July , the Conmaicne , Esperance and \wughighlight{Sappallina} became the first to find the existence of a feature in the legality of the construction of the new site . The temple had a little capacity and a much more expensive and rural one . The property was constructed across the river as a result of news of the ongoing criminal movement , and a major coastal post and a major movement of their range . In addition , the government of the United States and the west , and the Andersons were in charge of the building and \wughighlight{commentelist} of the nearby Fort Lofty Dam in the area . The bodies were based on the land and medical activities in the city , as well as the larger area of the city ; this was even the first of the largest and most expensive and significant issues of the world .

}

\subsection{$|\mathcal{V}| = 60000, T = 1.0$}

{\footnotesize\sffamily\color{gray}

The Songyue Wakō ( Sargasso Away ) of Kijirō Ola , the recruited daughter of Ka castigada ( " Yellow Cross , " ) references the royal same fortification the ' Footloose Festival ' . He even is the first person to reach its number , which was hospitalized during the visitor phase 's birth and appeared at the J @-@ eighties . On television treatment Kirk Williams published Selected reports with his father Bernard I of the United States that 's Shorea Promota using a plane reactor that wrongly survives on the crypt at the double notes with Xavier beatings and adolescents on caravan or horseback gates . In the development of su rhyme , the patterns of which were shot in 1.f4 as vol . 1 White seeks to add the departure of the conventional yelling , which is the basis for the fiend .
}

\subsection{$|\mathcal{V}| = 40000, T = 0.75$}

{\footnotesize\sffamily\color{gray}
= = = \wughighlight{Councillading} = = =

Other improvements in the attack were to withdraws from the new \wughighlight{ciprechandred} by the \wughighlight{Semonism} . In the 1960s , the majority of the flocks were \wughighlight{dodged} and are thought to be in the task of having a wedding power . The first known in the early 1990s was a report on display and a collection of early names based on the narrative of the \wughighlight{Kasanayan} , which from the late 1960s to the 19th century were made by a Dutch and American writer , \wughighlight{Astrace} \wughighlight{Barves} . The last larger , \wughighlight{wheelers} , piece of \wughighlight{muter} that was used to mark \wughighlight{Orremont} of the Old Pine Church was subsequently found in the National Register of Historic Places .
}

\subsection{$|\mathcal{V}| = 20000, T = 0.75$}

{\footnotesize\sffamily\color{gray}

The first major disaster on the island was the \wughighlight{Puree} of the \wughighlight{Greetistant} , which was the first \wughighlight{tightpower} the \wughighlight{Sconforms} of their lives , and the \wughighlight{noughouse} of chip and \wughighlight{woofbather} . \wughighlight{Ranching} later became \wughighlight{polluting} . The \wughighlight{senachine} were made by the efforts of Dr. \wughighlight{Berka} \wughighlight{Merroinan} , who had been also to \wughighlight{stair} for one of the previous cities .

= = = \wughighlight{Sinical} = = =

Further south of the population , the excavated area , though some of the " most important @-@ known \wughighlight{conventive} of the life of a more important mother " , was the \wughighlight{substation} of \wughighlight{reinstate} , the first U.S. state of the Stone .

In the early 20th century the American government passed a mission to expand the work of the building and the construction of the new collection . In the early 19th century , the \wughighlight{Synchtopic} was more prominent than the explorers in the region , and the young German \wughighlight{mainteness} , who were often referred to as the " \wughighlight{Poneoporacea} \wughighlight{Bortn} " , were persuaded to sit on their own \wughighlight{braces} . They had \wughighlight{sanited} with all a new \wughighlight{confistence} , and the religious \wughighlight{ottended} led to the arrival of the \wughighlight{Rakrako} family \wughighlight{Dombard} . Following the death of Edward McCartney in \wughighlight{1060} , the new definition was transferred to the \wughighlight{WDIC} of \wughighlight{Fullett} . The new construction was begun in \wughighlight{1136} . Several years later , the fundamental interest of the site was to be used after its death , where the \wughighlight{signate} were still to be built .
}

\subsection{$|\mathcal{V}| = 5000, T = 0.75$}

{\footnotesize\sffamily\color{gray}
The \wughighlight{Evang} was the first and first \wughighlight{anthropia} to be held in \wughighlight{unorganism} . \wughighlight{Wiziges} were some of the first in the region , and the \wughighlight{remarkable} was a \wughighlight{owline} . The \wughighlight{sale} was \wughighlight{sittling} to \wughighlight{timber} \wughighlight{Robert} , a independent family , and a \wughighlight{batting} of prime minister \wughighlight{Gillita} \wughighlight{Braze} , who was \wughighlight{noil} to the \wughighlight{Lokersberms} in the \wughighlight{frankfully} of the \wughighlight{playe} ' \wughighlight{undertook} \wughighlight{philippines} . The \wughighlight{particles} of the \wughighlight{tranquisit} \wughighlight{Full} were influenced by the \wughighlight{companies} and \wughighlight{intercourse} of the \wughighlight{Pallitz} , but the \wughighlight{comparison} of the \wughighlight{corishing} 's application was not known .
}

\end{document}